\pdfoutput=1
\documentclass[11pt]{article}

\usepackage{ACL2023}

\usepackage{times}
\usepackage{latexsym}

\usepackage[T1]{fontenc}

\usepackage[utf8]{inputenc}

\usepackage{microtype}

\usepackage{inconsolata}

\usepackage{hyperref}
\usepackage{graphicx}
    \graphicspath{{./images}}
\usepackage{xcolor}
\usepackage[english]{babel}
\usepackage{csquotes}
\usepackage{physics}
\usepackage{amsmath, amsfonts,amssymb}
\usepackage{numprint}
\usepackage{booktabs}
\usepackage[flushleft]{threeparttable}
\usepackage{caption}
\usepackage{subcaption}

\newcommand{\note}[1]{{\color{blue} \textbf{#1}}}
\newcommand{\probdist}{p}
\newcommand{\traindata}{\mathcal{D}_T}
\newcommand{\valdata}{\mathcal{D}_V}

\title{Entangling Machine Learning with Quantum Tensor Networks}

\author{Constantijn van der Poel \\
  Sandbox AQ \\
  \texttt{constantijn@sandboxaq.com} \\\And
  Dan Zhao \\
  New York University \\
  Sandbox AQ \\
  \texttt{dan.zhao@sandboxaq.com} \\}

\begin{document}
\maketitle

\begin{abstract}
This paper examines the use of tensor networks, which can efficiently represent high-dimensional quantum states, in language modeling.
It is a distillation and continuation of the work done in \cite{my_thesis}.
To do so, we will abstract the problem down to modeling Motzkin spin chains, which exhibit long-range correlations reminiscent of those found in language.
The Matrix Product State (MPS), also known as the tensor train, has a bond dimension which scales as the length of the sequence it models.
To combat this, we use the factored core MPS, whose bond dimension scales sub-linearly.
We find that the tensor models reach near perfect classifying ability, and maintain a stable level of performance as the number of valid training examples is decreased.
\end{abstract}

\section{Introduction}
Quantum tensor networks have achieved great success in many-body quantum systems, due to their ability to efficiently represent high-dimensional state vectors.
Furthermore, there has been a growth of interest in their application to machine learning problems \cite{davidSupTensNet}.
In this paper, we are interested in exploring the application of tensor networks to learn the Motzkin spin chain dataset.
We run experiments to learn this dataset via Stochastic Gradient Descent (SGD), to demonstrate a proof of concept of a Matrix Product State (MPS), and a novel model we call the \textit{factored core} MPS.

This paper builds upon the work done for the doctoral thesis of Tai-Danae Bradley \cite{tai_thesis}.
This in turn was built on the works of John Terilla et al. [\cite{Pestun2017LanguageAA}, \cite{Stokes_2019}, \cite{Miller_2020}].
The idea of using tensor networks to do machine learning can be seen in \textit{Exponential Machines} \cite{exp_mach}.
Here, Novikov, Trofimov, and Oseledets introduce the Tensor Train (TT) as a method of factorizing a linear regression model.
They also develop a stochastic Riemannian optimization procedure to optimize their model.
Concurrently, the work of Stoudenmire and Schwab \cite{schwab_tn} used tensor networks for image processing.
They used the DMRG algorithm \cite{dmrg} instead of the stochastic Riemannian optimization.
The advantage of using DMRG is that it automatically adjusts the bond dimensions of the underlying model, but it is computationally expensive.
Other works involving tensor networks include \cite{Liu_2019}, which trains a two-dimensional tensor network for image recognition; \cite{guifre}, which investigates entanglement structures with image classification; and \cite{Tangpanitanon_2022}, which uses the tensor networks to analyze Recurrent Arithmetic Circuits (RACs).

\section{Theoretical Background}
We will focus on sequences of finite fixed length, where each element of the sequence comes from the set $\mathcal{V}$, the \textit{vocabulary}.
We denote its cardinality as $|\mathcal{V}|=V$.
We will follow the procedure developed in \cite{Qstates_seq}.

Assume $s\in \mathcal{S}_n$ is a sequence in the set of length-$n$ sequences built from elements from $\mathcal{V}$.
Assume furthermore that $\probdist(s)$ is a probability distribution on the sequences; this is what we want our model to learn.
We will map each sequence $s$ in $\mathcal{S}_n$ to an orthonormal basis vector in the free vector space $\ket{s}\in\mathbb{C}^{\mathcal{S}_n}$.
Our model, $\ket{\psi}$, aims to learn $\probdist(s)$ such that it satisfies the \textit{Born Rule}:
\begin{equation}\label{eq:born_rule}
	\abs{\braket{s}{\psi}}^2=\probdist(s)
\end{equation}

To do so, we will decompose the input sequences into combinations of the constituent elements.
As we can decompose the sequences, so too can we decompose the vectors they map to.
We do this using the \textit{tensor product}.
Let us begin with a simple decomposition.
Imagine that we decompose our set of sequences, $\mathcal{S}_n$, into a Cartesian product of sets, $\mathcal{X}$ and $\mathcal{Y}$, which we call \textit{prefixes} and \textit{suffixes}, using the terminology of \textit{Definition 3.1} of \cite{tai_thesis}.
Let both $\mathcal{X}$ and $\mathcal{Y}$ have fixed subsequence lengths; therefore, each $s\in\mathcal{S}_n$ can be unambiguously split into $s=x+y$, for $x\in\mathcal{X}$ and $y\in\mathcal{Y}$.

\begin{figure}
    \centering
    \includegraphics[width=\linewidth]{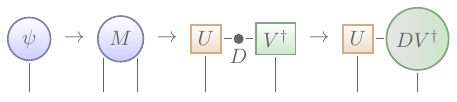}
    \caption{\label{fig:vec_decomp}Vector decomposition, singular value decomposition, absorbing diagonal matrix.}
\end{figure}

We now perform this decomposition with tensor networks.
A basic understanding of tensor networks and their diagrams is assumed, but we will review core concepts.
With tensor networks, each shape is a tensor, and the number of legs it has indicates its number of indices.
Therefore, a vector has one leg, and a linear map (a matrix) has two.
Diagrammatically (as in Figure (\ref{fig:vec_decomp})), we can imagine the decomposition as first reshaping our model, $\ket{\psi}$, from a vector into a linear map.
This is programmatically equivalent to reshaping a vector array into an order-2 matrix, $M$.

Now, we can apply the Singular Value Decomposition (SVD) to rewrite the matrix as the product of a unitary matrix, a diagonal matrix (with zeros for padding if $M$ is not square, to get the shapes to line up), and the conjugate transpose of another unitary matrix: $M=UDV^\dagger$.
We can then absorb $D$ in $U$ or $V^\dagger$, see Figure (\ref{fig:vec_decomp}), where $D$ is absorbed into $V^\dagger$).
We can think of the decomposed vector as a map between the subspaces $X$ and $Y$.
As such, the internal bond in the rightmost diagram of Figure (\ref{fig:vec_decomp}) can be thought of as a pathway to transport of information.

Extending this principle to decompose the entire sequence, we have
\begin{equation}
        \ket{s} = \ket{s_0} \otimes\ket{s_1} \otimes \ldots \otimes\ket{s_n} \in\mathbb{R}^{\mathcal{S}_n}.
\end{equation}
Note that we have switched to using real numbers.
This is done primarily to simplify the programming.
We will find in our experiments that the models perform well without needing to use complex numbers.
Model performance when using complex parameters is left as an avenue for future work.
For each vector, $\ket{s_i}$, in the tensor product, we have a corresponding tensor in our tensor network.
The $\ket{s_i}$ are the standard basis vectors of $\mathbb{R}^\mathcal{V}$, where $\mathcal{V}$ is our vocabulary.

\subsection{Dense Core MPS}\label{sec:dense_core}
\begin{figure}
    \centering
    \includegraphics{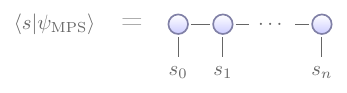}
    \caption{Ket and diagrammatic notation.}
    \label{fig:ket_mps}
\end{figure}

We will denote our specific model, a \textit{matrix product state} approximation of the desired (and unknown) state $\ket{\psi}$, as $\ket{\psi_\text{MPS}}$:
If we wish to contract the input sequence with a model, we can write it in either bra-ket notation, or diagrammatically, as shown in Figure (\ref{fig:ket_mps}).
Assume also that internal bonds have dimension $\chi$, and external (physical) bonds have dimension $V$.

\begin{figure}
    \centering
    \includegraphics{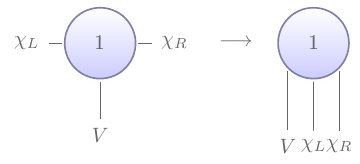}
    \caption{Index ordering.}
    \label{fig:ind_ord}
\end{figure}

For use with numpy-like code (e.g. Jax), it is useful to create diagrams that explicitly indicate the ordering of the indices, which will be relevant with reshaping operations during the contraction process.
As an example, examine tensor $1$ from Figure (\ref{fig:ket_mps}).
As a convention, we take the first index to be the external index, then take the internal indices from left to right.
Let us label the tensors by their position in the chain.
The legs are labeled with the size of their dimensions, as shown in Figure (\ref{fig:ind_ord}).

\subsection{MPS Norm} \label{sec:mps_norm}
\begin{figure*}
    \centering
    \includegraphics{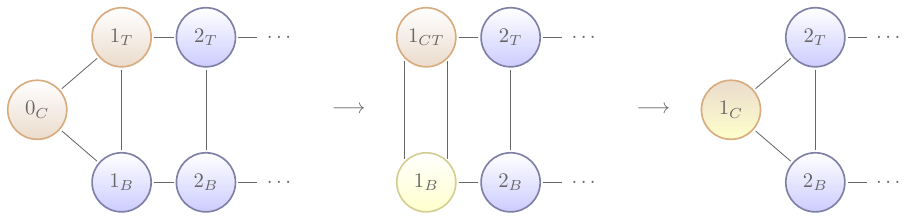}
    \caption{Contraction loop.}
    \label{fig:mps_loop}
\end{figure*}

In order to interpret the contraction of an input with our model as a probability, we must ensure the model is normalized: $\braket{\psi_\text{MPS}}=1$.
We therefore need an algorithm to calculate the contraction of the MPS with itself.
This needs to be done carefully, to ensure the computational complexity does not exceed $\mathcal{O}(\chi^3)$.
We show diagrams for the first of these steps for the sake of concreteness; the position-index of the tensors can be incremented without loss of generality.
These steps, show in Figure (\ref{fig:mps_loop}), both have a cost of $V\chi^3$.

\begin{figure}
    \centering
    \includegraphics{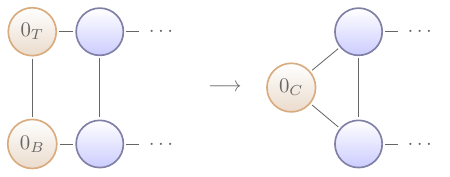}
    \caption{Cap contraction.}
    \label{fig:mps_cap}
\end{figure}

Let us dive into the details.
We start by contracting the tensors at one end over their external dimension, to create what we shall call a \textquote{cap}.
We use subscripts to denote the copy of tensor 0 on the top ($T$) and bottom ($B$), as well as the resulting cap ($C$). Again, the tensors are labeled by their position in the chain, see Figure (\ref{fig:mps_cap}).

\begin{figure}
    \centering
    \includegraphics{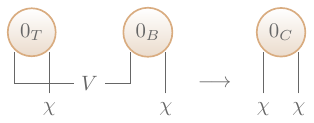}
    \caption{Cap contraction, index ordered.}
    \label{fig:mps_cap_ind}
\end{figure}

This contraction requires a computational cost of $V\chi^2$ operations.
The index-ordered diagram (Figure (\ref{fig:mps_cap_ind})) shows that we transpose the top copy of tensor 0, as can be seen by its crossing legs.
Since we are multiplying the tensor at position 0 by its transpose, shown in Figure (\ref{fig:mps_cap_ind}), the resulting matrix will be symmetric.

\begin{figure}
    \centering
    \begin{subfigure}[b]{0.4\textwidth}
        \centering
        \includegraphics{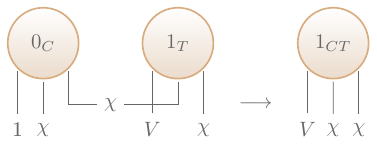}
        \caption{Contracting in the cap,}
        \label{subfig:mps_loop_cap_in}
    \end{subfigure}
    \hfill
    \begin{subfigure}[b]{0.4\textwidth}
        \centering
        \includegraphics[width=\linewidth]{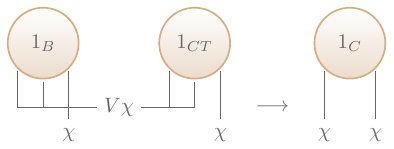}
        \caption{Creating the next cap;}
        \label{subfig:mps_loop_new_cap}
    \end{subfigure}
       \caption{Index ordered diagrams for the contraction loop.}
       \label{fig:mps_loop_ind}
\end{figure}

From here, we can follow a pattern of two steps down the chain, namely, contracting the cap with the top tensor, then creating a new cap.
The first step in this two-step loop can be done with a straightforward batched matrix multiplication (assuming an implicit extra leading dimension of one for the cap).
Note that the rightmost index of the cap technically corresponds to the bottom bond instead of connecting to the top,
but we take advantage of the symmetry of the cap, and avoid doing a transpose operation.
This is shown in Figure (\ref{subfig:mps_loop_cap_in}).
The second step requires the merging of dimensions in the relevant tensors, as shown in Figure (\ref{subfig:mps_loop_new_cap}).
After this, the second step is just a matrix multiplication (where we have also transposed the reshaped bottom tensor), bringing us back to a symmetric cap.

\begin{figure}
    \centering
    \begin{subfigure}[b]{0.4\textwidth}
        \centering
        \includegraphics[width=\linewidth]{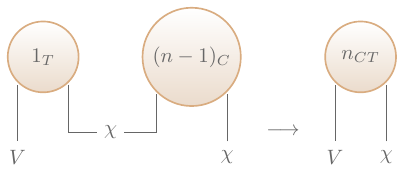}
        \caption{Contracting in the final cap,}
        \label{subfig:fin_cap}
    \end{subfigure}
    \hfill
    \begin{subfigure}[b]{0.4\textwidth}
        \centering
        \includegraphics{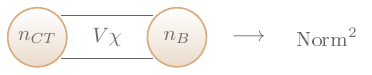}
        \caption{Getting the norm;}
        \label{subfig:norm_sq}
    \end{subfigure}
       \caption{Final steps in MPS norm calculation.}
       \label{fig:mps_loop_final}
\end{figure}

This loop will eventually bring us to the final cap.
From here, the contraction proceeds almost like the above loop, but with one less index, see Figure (\ref{subfig:fin_cap}).
To contract the final two tensors, $n_{CT}$ and $n_B$, we flatten them and do standard vector multiplication (the inner product) to yield the final norm squared value, shown in Figure (\ref{subfig:norm_sq}).

\section{Motzkin Spin Chains\label{sec:motzkin}}
In this section, we examine the dataset we will be using to conduct our experiments.
The so called Motzkin numbers occur in a multitude of manifestations, as demonstrated in \cite{motzkin_numbers}.
Here, we will be interested in Motzkin numbers as they count the number of possible Motzkin walks of length $n$.

\begin{figure}
    \centering
    \includegraphics{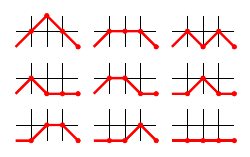}
    \caption{All valid Motzkin walks of length four for spin $s=1$ \cite{motzkin_pic}.}
    \label{fig:motzkin_walks}
\end{figure}

A Motzkin walk can be constructed for a spin chain of length $n$ and spin $s\in\mathbb{N}$.
Recall that, for spin $s$, there are $2s+1$ values for the magnitude along an axis (typically, $z$).
We consider the middle value (zero) to be a flat step (+1,0), positive values to be an upward step (+1,+1; we differentiate these with a label, such as color), and the negative values are corresponding downward steps (+1,-1).
We will focus our attention on Motzkin walks with spin $s=1$.
A valid Motzkin walk is one that starts at (0,0) and ends at (0,0), without ever dropping below $y=0$.
All valid Motzkin walks of length four for spin $s=1$, are shown in Figure (\ref{fig:motzkin_walks}).

\begin{figure}
    \centering
    \includegraphics[width=\linewidth]{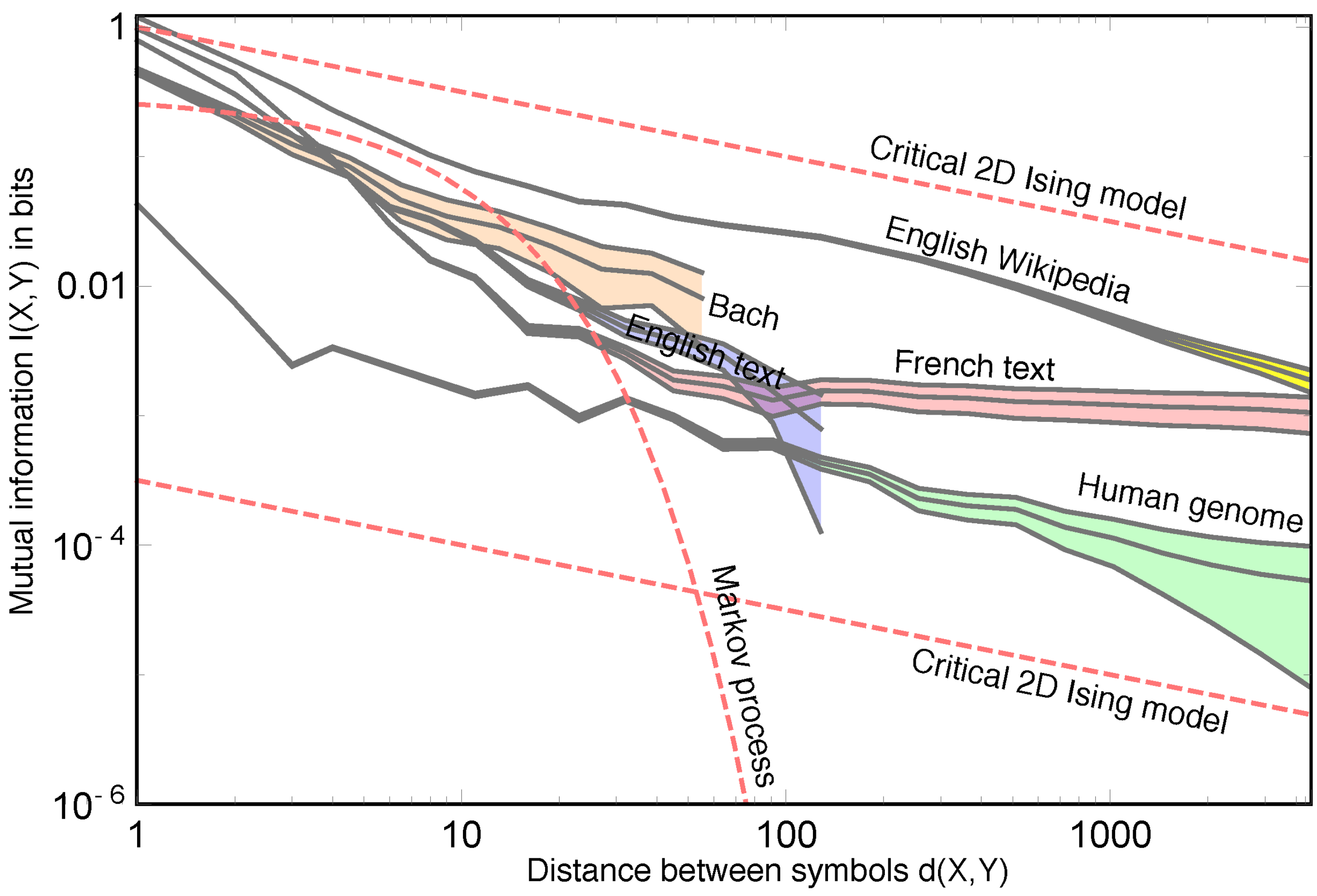}
    \caption{\label{fig:crit_lang}Mutual information is plotted as a function of token distance in natural language, music, and Markov processes \cite{crit_formal_lang}.}
\end{figure}

Consider Figure (1) of \cite{crit_formal_lang}, shown in Figure (\ref{fig:crit_lang}).
Here, we see that many sequences of interest, such as language, music, and DNA, have long-range, \textit{power-law} scaling correlations in the mutual information as a function of token separation distance.
We can calculate this plot for the Motzkin chains as well:
\begin{figure}
    \centering
    \includegraphics[width=\linewidth]{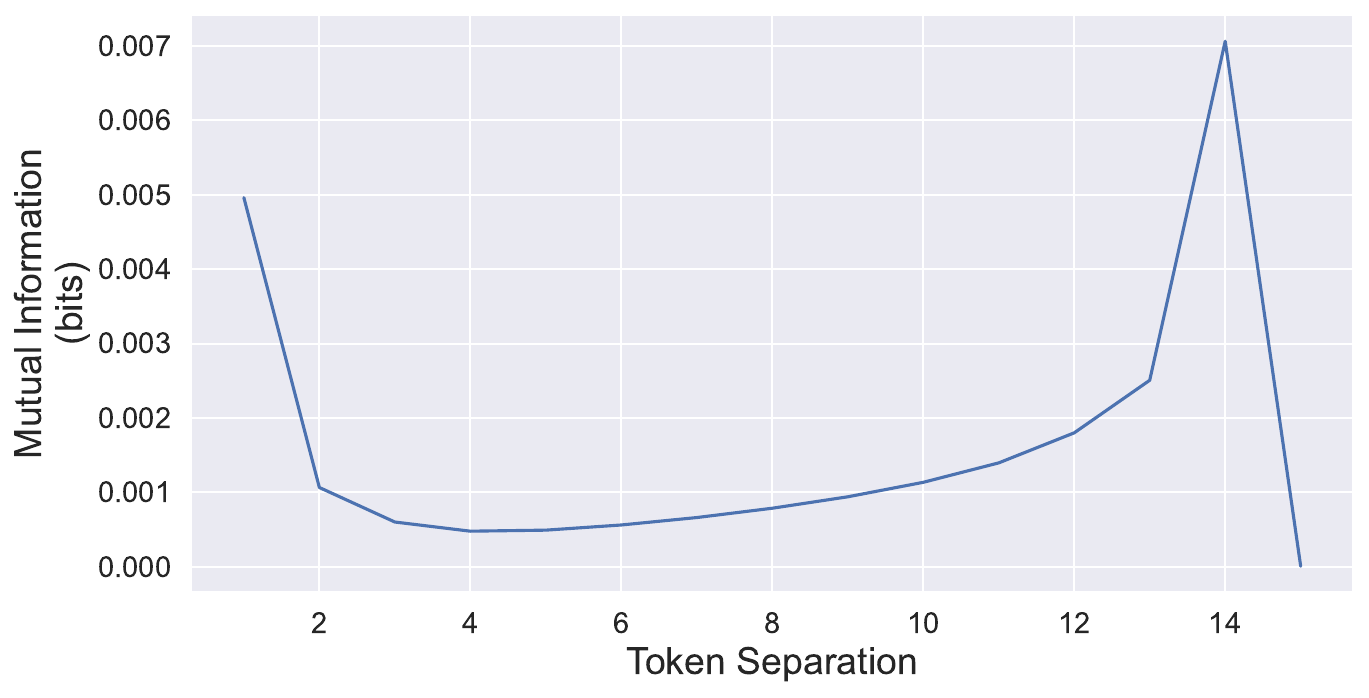}
    \caption{\label{fig:mutual_info}Mutual information vs. token distance for Motzkin sequences of length 16.}
\end{figure}
The sharp peak for high token separation is likely due to boundary effects.
While the insides of Motzkin sequences are fairly flexible, the ends are correlated, since they contain matching open and close parentheses that are needed to complete the remaining pairs.

These Motzkin spin chains were first studied as a system of interest to the study of quantum many-body systems (QMBSs) \cite{crit_spin1_chain}.
In \cite{crit_spin1_chain}, the authors show that the entanglement entropy between half-chain subsystems scales as $\ln(n)$, where $n$ is the chain length.
For an MPS, the area law indicates that the entanglement entropy scales as $\ln(\chi)$. 
Thus, the bond dimension would have to scale linearly with system size with the MPS: $\chi\sim n$.
The number of operations to contract two inner cores (neither first nor last in the sequence) would scale as $n^3$, which quickly becomes infeasible on contemporary hardware for modest sequence lengths in the hundreds.

\subsection{Factored Core MPS}
\begin{figure}
    \centering
    \includegraphics{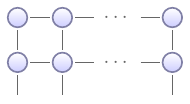}
    \caption{Factored core model, $h=2$.}
    \label{fig:fact_mps}
\end{figure}

This motivates the use of a \textit{factored core} MPS.
For the factored core, we can think of the dense cores as turned into vertical products of cores.
Let $h$ represent the height of the factored core model: the number of tensors in the vertical direction. In Figure (\ref{fig:fact_mps}), $h=2$.
We now differentiate between vertical and horizontal internal bonds, whose dimensions are denoted by $\chi_v$ and $\chi_h$. 
The subscript $h$ in $\chi_h$ merely indicates that we are using the \textit{horizontal} bond dimension, and is unrelated to the factored core's height.

\begin{figure}
    \centering
    \includegraphics{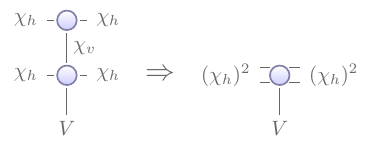}
    \caption{\label{fig:factored_core}Factored core contraction, $h=2$.}
\end{figure}

When contracting the factored core model, we contract the cores vertically first, as shown in Figure (\ref{fig:factored_core}).
This yields a network with the same topology as the dense cores, so the dense algorithms in sections (\ref{sec:dense_core}) and (\ref{sec:mps_norm}) apply from this point on.
In our experiment, we will simply merge the internal bond dimensions of the factored cores, such that the resulting effective dense core has bond dimension ${\chi_h}^h$.

\begin{figure}
    \centering
    \includegraphics{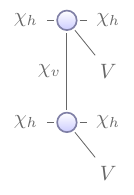}
    \caption{\label{fig:skip_core}Factored core with skip connections}
\end{figure}

In the factored core shown in Figure (\ref{fig:factored_core}), information must flow from the physical index up through the core.
As the core height increases, we expect the information from the physical index to degrade as it moves up the subcores.
This can be addressed by constructing a \textit{skip connection}: where there is a copy of the physical index on each subcore.
A diagram is shown in Figure (\ref{fig:skip_core}).

\begin{figure}
    \centering
    \includegraphics{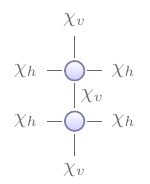}
    \caption{\label{fig:factored_core_scaling}Most computationally intensive possible factored core sub-contraction.}
\end{figure}

We claim that their contractions scale at most as
\begin{equation}\label{eq:fact_scaling}
	{\chi_h}^4{\chi_v}^3 \sim \ln^3(n)
\end{equation}
The most computationally intensive step comes from the contraction shown in Figure (\ref{fig:factored_core_scaling}), which occurs for factored cores of height of at least four.
An analytic MERA solution exists for the Motzkin spin chain system \cite{Alexander_2021}.
We imagine a factored core network as loosely topologically encompassing the MERA network.
Since the MERA solution has $h\sim\ln(n)$, we assume the same for the factored cores.
A factored core equivalent of a dense core solution would have ${\chi_h}^h=\chi$.
Then, we expect
\begin{equation}
	\chi_h \sim \chi^{1/h} \sim n^{1/\ln(n)} = e.
\end{equation}
Analogously to $\chi\sim n$ for the MPS (imagine a factored core as a vertical MPS), we expect our vertical bond dimension to scale as the height (\enquote{vertical length}), so
\begin{equation}
	\chi_v\sim h\sim \ln(n).
\end{equation}
With these, we arrive at Eq. (\ref{eq:fact_scaling}).

\section{Experiments}
\subsection{Neural Baseline}
Now that we have discussed the tensor network model, we want a baseline to compare it to.
We propose using a feed-forward neural architecture as a widely familiar benchmark.
This will give a familiar model for a wider AI audience to compare the tensor models' performance to.
We choose a Multi-Layer Perceptron (MLP) with a single hidden layer; a classic, and one that has shown to be performant if given a large enough hidden layer \cite{mlp_uni_approx}.
Each token in the input sequence, $s_i$, is first passed into an embedding layer, $\phi(\cdot)$, that maps it to a learnable vector in an embedding space of dimension $d_e$.
These are concatenated and passed into the hidden layer (which has $d_h$ neurons).
The neural model will output a single scalar, which is passed through a sigmoid nonlinearity.
The output will be denoted $\psi(s)$.

We will be comparing the tensor models to the neural model using a classification task.
Fundamentally, the two models are different; the MPS maps an input sequence to the probability of that sequence out of all sequences, while the neural model maps to its belief that the input is valid.
This is because the neural model has no form of normalization, while the tensor network inherently does.
For binary classification, the classification decision is made by comparing the scalar output to a threshold.
This threshold does not depend on normalization, so the task provides a level playing ground for the models.

\subsection{Training}
In order to test the efficacy of our models, we will run a series of experiments on the Motzkin dataset, using a sequence length of $n=16$.
Our goal is to learn the distribution, $\probdist(s)$, which is uniform over valid chains, and 0 elsewhere.
There are $\numprint{853467}$ valid Motzkin chains out of $3^{16}=\numprint{43046721}$ possible chains, which is a little less than 2\%.
The training dataset, $\traindata$, will consist of a random 25\% of the valid Motzkin chains.
To examine the tensor networks' performance during training, we use the set of all valid Motzkin chains; we will call this the validation set, $\valdata$.
We are interested in seeing if the fraction of probability mass put on the training data versus on all valid chains equals the fraction of training data to all valid chains.
We use this as a measure of generalization: that the model is properly extrapolating the pattern of the training data, and not just placing all probability mass on the training data.
The latter would also cause the sum over the probabilities of all valid chains to equal one.

We will point out that, usually, the training data is purposefully \textbf{excluded} from the validation set.
In this case, we include it to calculate the \textit{true} total probability mass the model puts on all valid chains, not just a statistical estimate.
We are able to do this since we are working with a dataset for which we can feasibly calculate the output of \textit{every} valid sequence.
Because we are able to calculate the true generalization, there should be no notion of data contamination.
This is, of course, not the case with \enquote{real world} data, and is a luxury that lets us focus on the science instead of copious hyperparameter tuning.

We use the binary cross-entropy loss:
\begin{gather}\label{eq:loss}
	\mathcal{L} = -\mathbb{E}_{\traindata}\left[ y\mathop{lp}(x) + (1-y)\ln\big(1-e^{\mathop{lp}(x)}\big) \right],\\
	\mathop{lp}(x) \equiv 2\ln{\braket{\psi}{x}}-\ln\braket{\psi},
\end{gather}
where $\traindata$ is comprised of inputs and binary labels, $(x,y)$, and $\mathop{lp}(x)$ is the log of the probability of sequence $x$, rewritten from Eq. (\ref{eq:born_rule}).

For a validation epoch, we contract the model with all $x\in\valdata$.
Summing all these probabilities will yield the probability of the model's outputting a valid chain:
\begin{equation}\label{eq:sigmav}
	\Sigma_V \equiv \sum_{x\in\valdata} \frac{\braket{\psi}{x}\braket{x}{\psi}}{\braket{\psi}}
\end{equation}
The complement of this probability will of course be the model's probability of outputting an invalid chain, which should go to zero.

A metric commonly used in language modeling tasks is \textit{perplexity}.
The perplexity of a statistical model, $\psi$, is defined as
\begin{equation}
	\mathop{PP}(\psi) \equiv \exp{H(\psi)}
\end{equation}
Here, $\mathop{H}(\psi)$ is the entropy of the model.
Note that the output of the model needs to be the probability of the input sequence, not the probability that the input is valid!
Calculating the perplexity for neural models can get a bit hazy, with conventions varying.
Because of this, we will stick to using $\Sigma_V$ as our primary metric of interest.

\section{Tensor Network Initialization}
\subsection{Dense Core Initialization}
We would like to initialize the model such that it generates a uniform (the maximum entropy) distribution, plus some noise, over the space of sequences it operates on.
To do this, we initialize each core (except the left-/right-most cores, which we shall call the outer cores) to be an identity matrix (plus some noise) for each slice along the physical index.
We denote this with $I_\chi$: the identity matrix of rank $\chi$.
We initialize the outer cores to be ones (again, for each slice along the physical index), scaled by a factor of $1/\sqrt{\chi}$, plus some noise.
We denote this with $\ket{1_\chi}$, a vector of length $\chi$ with all entries equal to one.
The leftmost core will be a bra instead of ket.
The motivation for this is, if we were to contract these ones vectors with the internal identities (ignoring the noise for a moment), we would end up with, after contracting with an arbitrary input string:
\begin{equation}
	\frac{1}{\sqrt{\chi}} \bra{1_\chi} I_\chi ... I_\chi \ket{1_\chi} \frac{1}{\sqrt{\chi}}
	= \frac{1}{\chi} \braket{1_\chi}{1_\chi} = \frac{\chi}{\chi} = 1
\end{equation}
After normalization, this will yield the uniform distribution, as desired.
We examine the initialization noise variance more in Sec. (\ref{sec:dense_init}), where we conclude that a solid initialization has the inner cores close to identity, to allow information to propagate more freely along the sequence in the early stages of training.

\subsection{\label{sec:fact_init}Factored Core Initialization}
We would like to initialize our factored cores to be equivalent after vertical contraction to a dense core initialized as in the previous section.
Recall Figure (\ref{fig:fact_mps}) for the dense equivalent of a factored core.
In some preliminary tests, we found that simply initializing the parameters from a uniform distribution (the default behavior for Flax layers) yielded a model unable to learn.
Thus, this initialization scheme was empirically found to be \textit{essential} to our model's learning. 
A deeper understanding of why this might be is left as an area for future work.

\begin{figure}
    \centering
    \includegraphics[width=\linewidth]{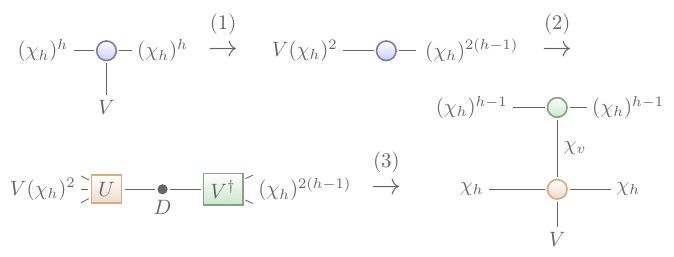}
    \caption{\label{fig:factorize}First step of iterating SVD to \enquote{factorize} a dense core.}
\end{figure}
We iterate application of the Singular Value Decomposition (SVD) to extend the core upward.
We call this process \enquote{factorizing} a dense core, the first step of which is shown in Figure (\ref{fig:factorize}).
We start with a dense core of bond dimension ${\chi_h}^h$, where $h$ is the desired height of the factorized core.

For step (1) in Figure (\ref{fig:factorize}), we reshape the dense core into an order-2 tensor, with the physical and one set of horizontal bond dimensions ($V{\chi_h}^2$) on one index, and the remaining sets of horizontal bond dimension on the other.
For outer cores, a set only has one power of $\chi_h$, not two, since the horizontal bonds are only on one side of the core.

For step (2), we perform the SVD.
Note that, unless the vertical bond dimension, $\chi_v$, equals the rank of the target tensor being decomposed ($D$), we will need to perform an extra step after the SVD.
If the desired $\chi_v$ is less than the target's rank, we need only discard the lowest singular values to trim down to number to $\chi_v$.
If the desired $\chi_v$ is greater than the target's rank, we add additional singular values.
The additional singular values are drawn from a uniform distribution; we default to a minimum value of 0.001 and a maximum of 0.01.

Step (3) concludes by reshaping the tensors.
After we have trimmed/extended $D$ to $\chi_v$, we reshape our bottom tensor, (labelled $U$), and absorb the diagonal $\chi_v$ tensor into the top tensor (labelled $V^\dagger$).
We can then reshape the top tensor, to finish the process, or iterate again to increase the core height further.
The result is a factored model that should evaluate to a properly initialized dense model.
Note that, whether or not the factored core has skip connections affects how close the factorized core is, after vertical contraction, to its dense \enquote{parent}.
The extra expressivity of having a physical index on each subcore, the case for skip connections, is needed to capture all of the parent core's information.

In addition to factorizing cores, SVD can also grant insight into a possible approximate contraction scheme.
This can be done by removing singular values from the diagonal matrix, $D$, of the SVD.
Say we have $M=UDV^\dagger$ in our SVD step, with $d_0$ the number of singular values in $D$.
If we remove singular values so this number reduces to $d_1$ (to yield a diagonal matrix $D'$), then the Frobenius norm of the difference between $M$ and $UD'V^\dagger=M'$ is
\begin{equation}
	\norm{M-M'} = \norm{M}\sqrt{\sum_{i>d_1}(s_i)^2},
\end{equation}
where $s_i$ are the singular values of $D$ (in this sum, they are the ones that have been removed to yield $D'$).
This is the \textit{restricted rank tensor approximation}; we can use it to quantify the loss in precision as we reduce the computational cost of the factored core complexities (recall these scale at most as ${\chi_h}^4{\chi_v}^3$).

\section{Hyperparameters}
In this section we will briefly discuss the experiments run to determine the models' sensitivities to various hyperparameters.
A more in-depth look will be reserved for the appendices.
There, we examine the effects of bond dimension (\ref{sec:bond_dim}), including the norm in the loss (\ref{sec:alpha_dep}), alternative norm calculations (\ref{sec:replace_norm}), and initialization noise variance (\ref{sec:dense_init}).
Based on the results of our hyperparameter tuning, we use a dense bond dimension of 8, and for the factored core model a horizontal dimension of 3 with a height of 2.
For the skip-factored core, we use a horizontal dimension of 2 and a height of 3, so its effective bond dimension is equal to the dense model's.
The model parameters used are collected in Table (\ref{tab:model_vars}), in the appendices.
We do not include a norm term in the loss, we calculate the full norm, instead of approximating, and use an initialization variance of 0.01

\subsection{Batch Size Dependence}\label{sec:bs}
We would like to highlight a particularly intriguing result we found when testing the effects of batch size on the dense core model.
We examine performance for batch sizes of powers of two between and including 2 and 1024.
\begin{figure}
    \centering
    \includegraphics[width=\linewidth]{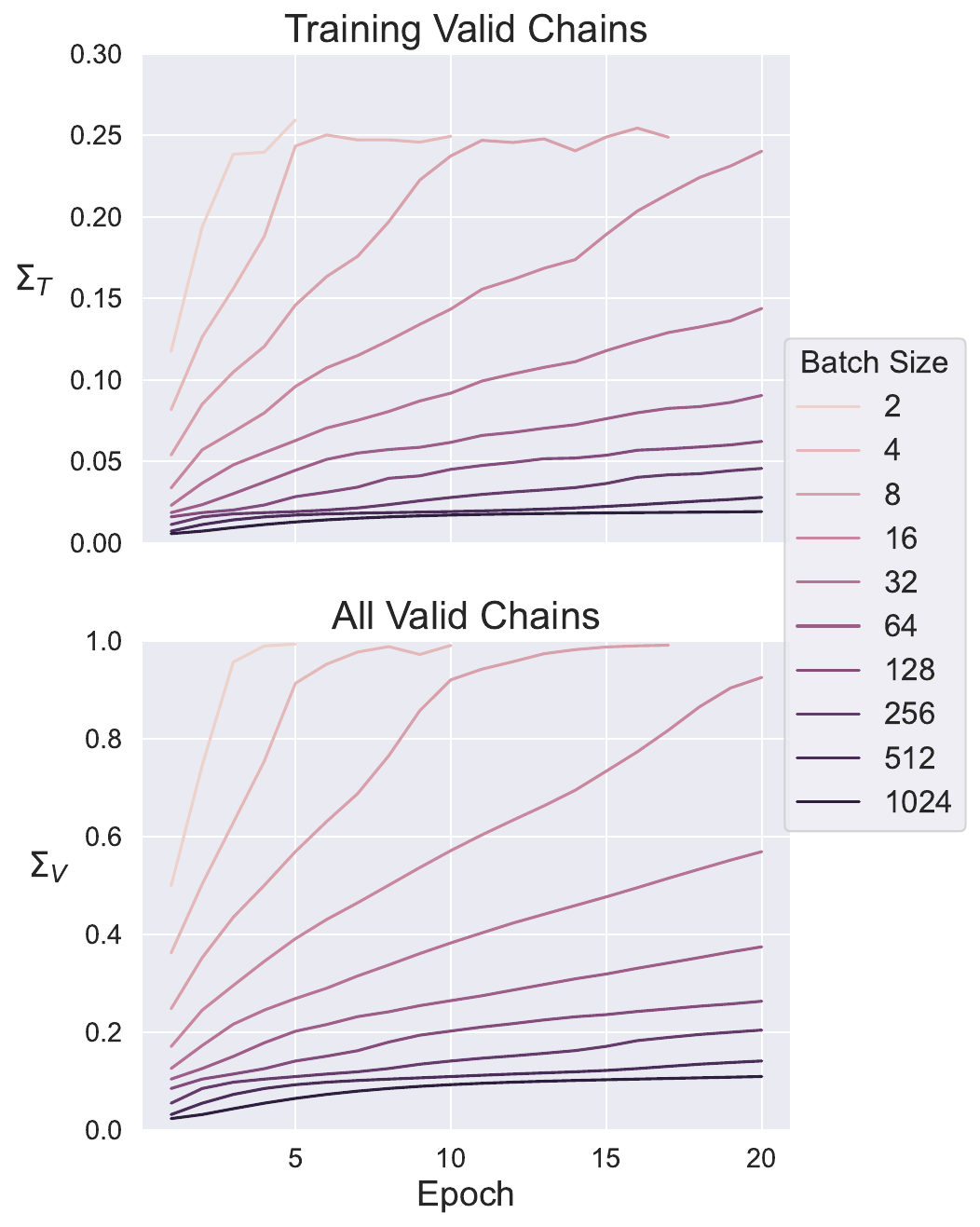}
    \caption{\label{fig:mps_bs}Batch size sweep for dense core model.}
\end{figure}
\begin{figure}
    \centering
    \includegraphics[width=\linewidth]{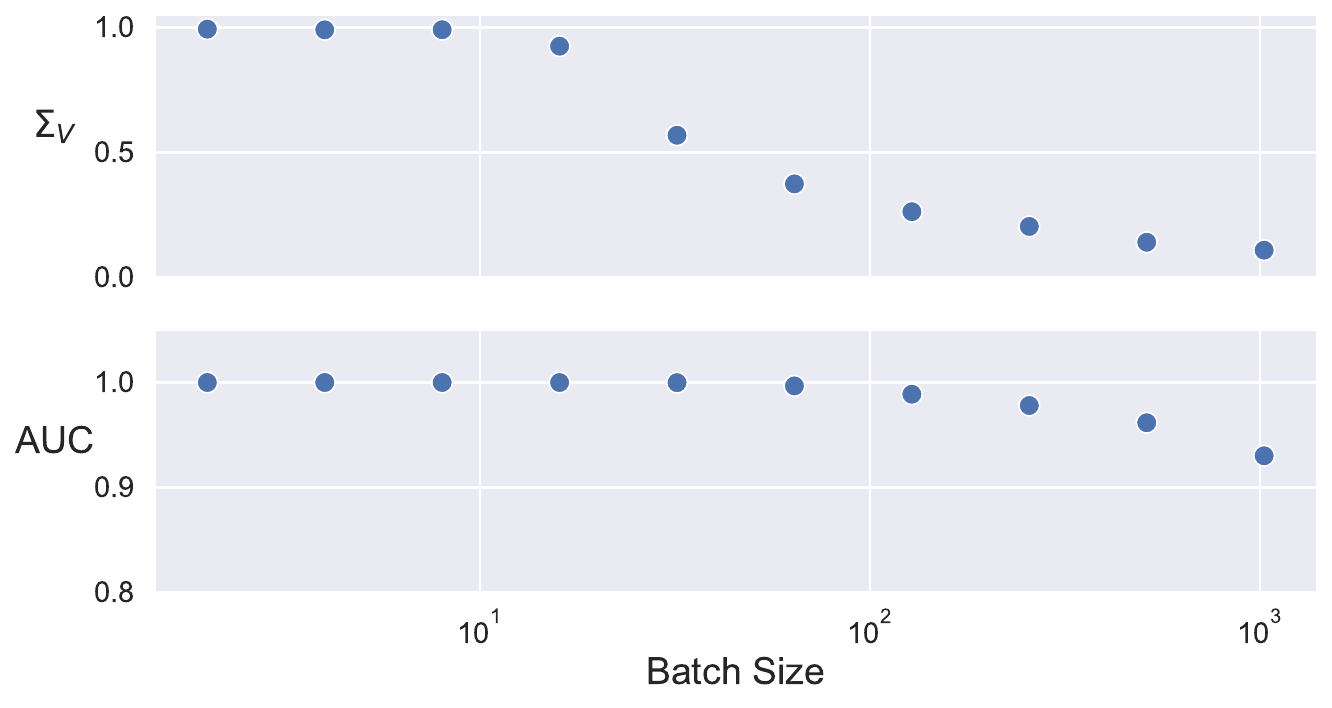}
    \caption{\label{fig:dense_bs_scat}ROC AUC metrics for batch size sweep for dense core model.}
\end{figure}
In Figure (\ref{fig:mps_bs}) and Figure (\ref{fig:dense_bs_scat}), we see that the dense core's performance degrades as the batch size increases.
Note in Figure (\ref{fig:dense_bs_scat}) that the $\Sigma_V$ metric begins falling off sooner and faster than the classifier (ROC AUC) metric.
We investigate the classifier performance for the neural and factored core models as well, shown in Figure (\ref{fig:bs_scat}).
\begin{figure*}
    \centering
    \includegraphics[width=0.75\textwidth]{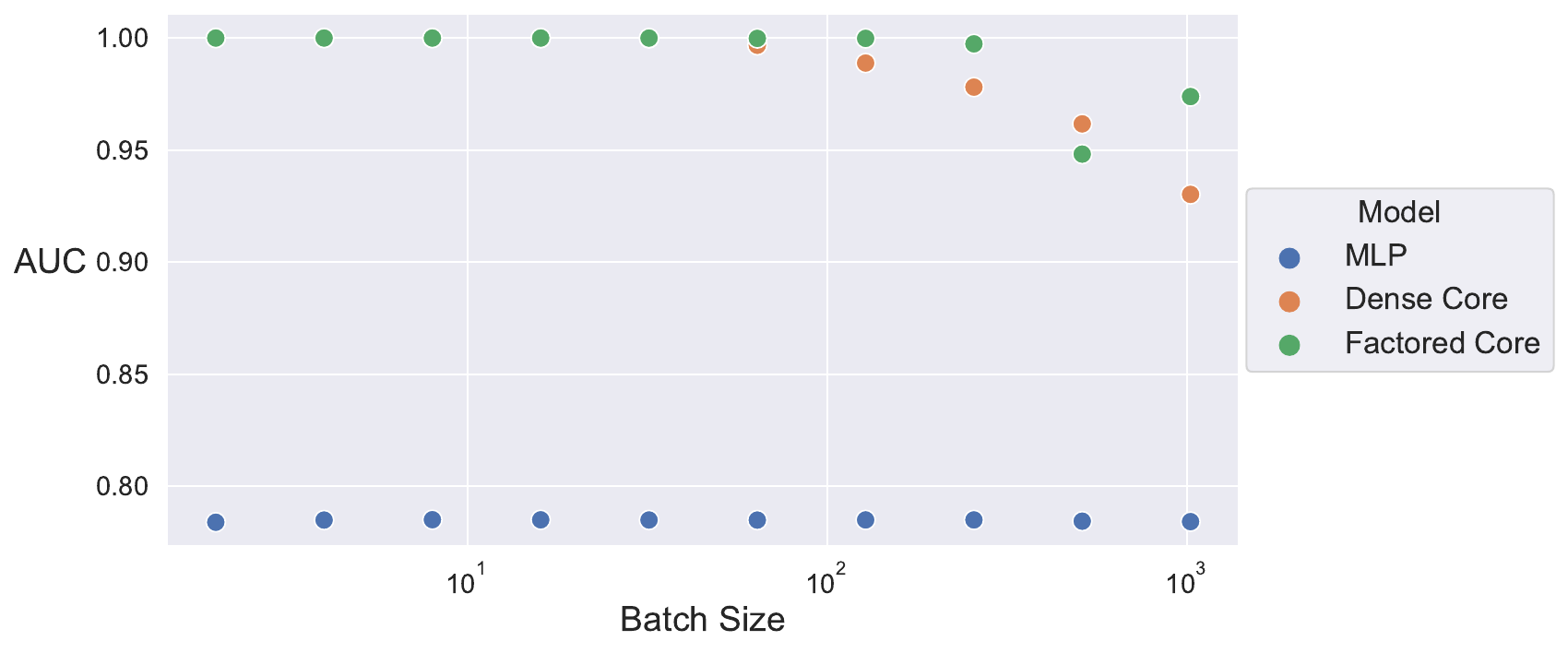}
    \caption{\label{fig:bs_scat}ROC AUC metrics vs batch size sweep for all models.}
\end{figure*}
Again, we see that the tensor networks' performance begins to drop for large batch sizes.
This is not the case for the neural model, which seems unaffected.
Currently, we are unable to determine why this is the case.
One possible explanation involves the fact that the tensor network models have a norm calculation to allow them to express explicit probabilities, instead of the neural model's \enquote{belief} probabilities.
We examined how removing and replacing the norm, in Sec. (\ref{sec:replace_norm}), affects learning, and since it reduces learning ability, we found this be an unlikely explanation.

We proceed to our primary experiment, training a binary classifier, and compare their performances using the ROC AUC metric.
Note that any AUC less than 0.5 will be replaced with $1-\text{AUC}$.
This is because, in that instance, the classifier is doing worse than random guessing.
As such, the opposite of its classification choices can be taken to yield better results.

\section{Classifier Task}
To begin, we use 10 random seeds to generate aggregate results.
We report averages to a precision determined by the standard deviation of the data; see Table (\ref{tab:10seed}).
\begin{table}
\begin{tabular}{lccc}\toprule
	Model & $\expval{\Sigma_T}$ & $\expval{\Sigma_V}$ & AUC \\\midrule
	\text{MLP} & - & - & 0.77(8) \\
	\text{Dense} & 0.252(8) & 0.993(1) & {\small0.999998(4)} \\
        \text{No-Skip} & 0.23(6) & 0.92(25) & 1.00(1) \\
        \text{Skip} & 0.2472(2) & {\small0.995415(2)} & {\small0.999999(2)} \\\bottomrule
\end{tabular}
\caption{\label{tab:10seed}10-seed evaluation metrics for the four model types.}
\end{table}

For example, we start with the neural model, which will serve as our baseline.
We run our Multi-Layer Perceptron (MLP) with embedding dimension $d_e=16$, and hidden layer size $d_h=256$.
Across ten random seeds, we see an average ROC AUC of 0.77 for the MLP, and a standard deviation of 0.08.
We will write this as 0.77(8) for compactness.
In this notation, we label our uncertainties (standard deviation) in parentheses, which correspond to the final digits of the measured average.

Our dense core tensor network model is run with a bond dimension of $\chi=8$.
In addition to achieving $\expval{\Sigma_T}$ and $\expval{\Sigma_V}$ near the ideal values of 0.25 and 1.0, respectively, the models effectively reach 1.0 on the ROC AUC metric.
Thus, the dense core model is able to effectively learn!
Moreover, it learns to classify much better than the neural baseline, and delivers consistent results across seeds.

Since we do not need to approximate the factored core's vertical contractions for the Motzkin dataset, we expect it to have similar results to the dense core model.
For the factored core tensor network without skip-connections, we use $h=2$, $\chi_h=3$, and $\chi_v=8$.
We see effectively the same performance, with the exception of an outlier datum, $\text{seed}=8$, that raises the variance.
We perform the same experiment with a skip-connection factored core, with $h=3$, $\chi_h=2$, and $\chi_v=4$.
Again, we see very similar performance between the tensor network models.

\begin{figure*}
    \centering
    \includegraphics[width=0.75\textwidth]{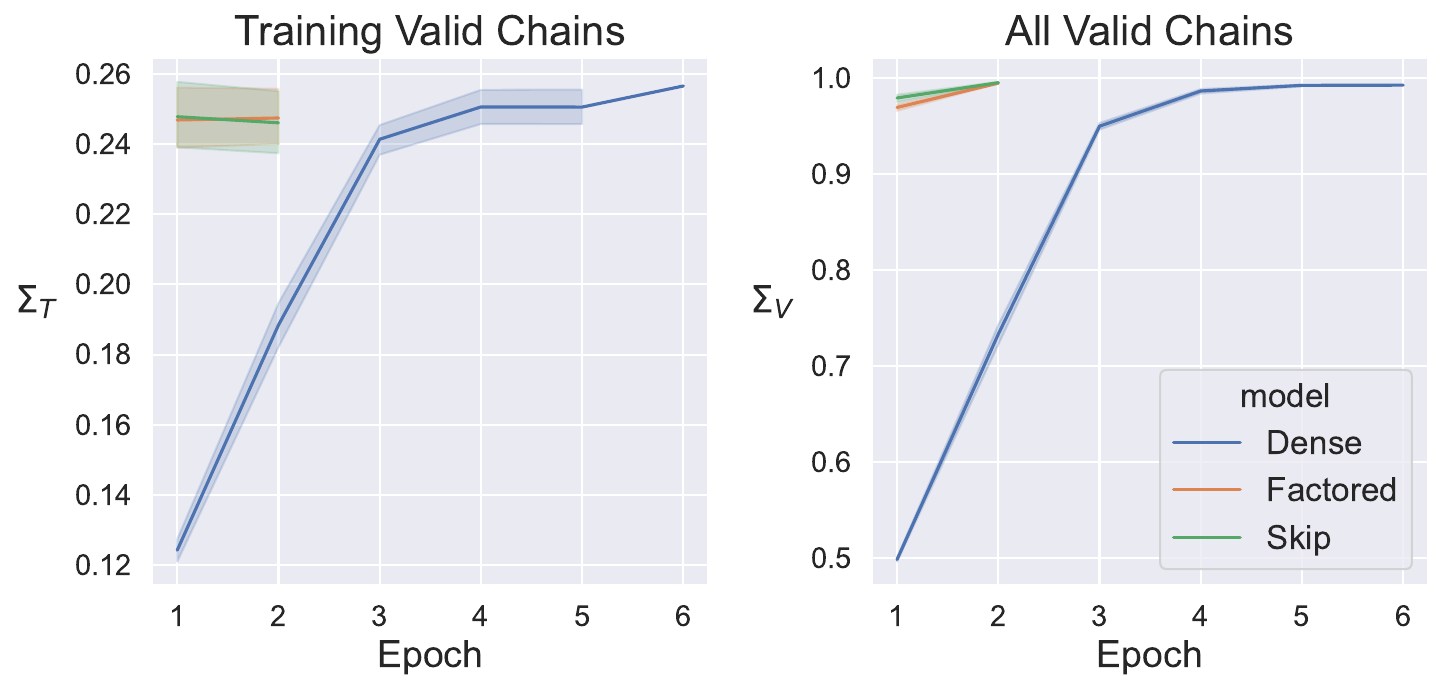}
    \caption{\label{fig:combo_9seed}Probability masses with 95\% confidence interval for 9 random seeds for tensor network models.}
\end{figure*}

Let us remove the seed that caused the outlier catastrophic forgetting for all models, and display both aggregate learning curves in Figure (\ref{fig:combo_9seed}).
For both models, we see tight 95\% confidence interval bands.
Without the outlier data, the factored core model is also quite consistent, and it is worth pointing out, though, that the factored models achieve $\Sigma_V\geq0.99$ in fewer epochs than their dense counterparts.

Consider the number of parameters each model uses, shown in Table (\ref{tab:core_sizes}).
Recall that the physical dimension is $V=3$ for the Motzkin dataset.
\begin{table}
\begin{tabular}{lccc}\toprule
	Position & Dense & Factored & Skip \\\midrule
	Outer    & 24	 & 96	    & 64 \\
	Inner	 & 192	 & 288	    & 128 \\
	Total ($n=16$) & 2736 & 4224 & 1920 \\\bottomrule
\end{tabular}
\caption{\label{tab:core_sizes}Core sizes for the tensor network models.}
\end{table}
Though for $n=16$ the factored core uses more parameters than the dense core after hyperparameter tuning, we see the superior scaling of the skip-factored core.
While the experiments intended to show that the factored core's performance wasn't too much worse than the dense core, we show that it in fact performs the same, even slightly better.
Not to mention, with an embedding dimension of $d_e=16$ and hidden layer size of $d_h=256$, the MLP uses \numprint{66352} parameters.
The tensor network models are able to capture the structure of the data using far fewer degrees of freedom, suggesting their architecture is a very good prior, for this dataset at least.

\section{Robustness to Sparse Valid Data}\label{sec:robustness}
Finally, we are interested in how our models respond to a lack of valid-chain signal.
In order to examine model robustness, let $\mu$ be the fraction of the training dataset that are valid Motzkin chains.
In the previous section, we restricted ourselves to $\mu=1.0$.
We will examine values of $\mu=1.0,0.75,0.5,0.25,0.1,0.01$, each with four random seeds.
The results are shown in Table (\ref{tab:mu}).
\begin{table}
\begin{tabular}{lllll}\toprule
	$\mu$ & MLP	 & Dense       \\\midrule
	1.0   & 0.81(4)  & 0.999997(4) \\
	0.75  & 0.74(3)  & 0.945(20)   \\
	0.5   & 0.70(3)  & 0.959(4)    \\
	0.25  & 0.61(6)  & 0.86(18)    \\
	0.1   & 0.58(7)  & 0.987(3)    \\
	0.01  & 0.57(4)  & 0.57(5)     \\\midrule
	$\mu$ & Factored & Skip \\\midrule
        1.0   & 0.999998(4) & 0.999998(2) \\
        0.75  & 0.939(16)   & 0.96(3) \\
        0.5   & 0.948(7)    & 0.97(3) \\
        0.25  & 0.940(25)   & 0.94(3) \\
        0.1   & 0.985(21)   & 0.9977(15) \\
        0.01  & 0.60(11)    & 0.521(2) \\\bottomrule
\end{tabular}
\caption{\label{tab:mu}ROC AUC vs. $\mu$ per model, averaged over four seeds.}
\end{table}
Again, we will start with our neural baseline.
The MLP's classification performance decreases monotonically with $\mu$; it performs better with a training dataset of purely valid data than a mix to contrast.
This is somewhat surprising, as conventional wisdom would predict that a balanced mix of valid and invalid samples would be ideal.
This could be due to the long-range structure in the data.
Coupled with the fact that valid Motzkin chains are rare among the set of possible chains, this suggests that each valid sample is quite important in helping the neural model learn the overall distribution.

We proceed to examine the $\mu$ robustness of the dense core model.
\begin{figure}
    \centering
    \includegraphics[width=\linewidth]{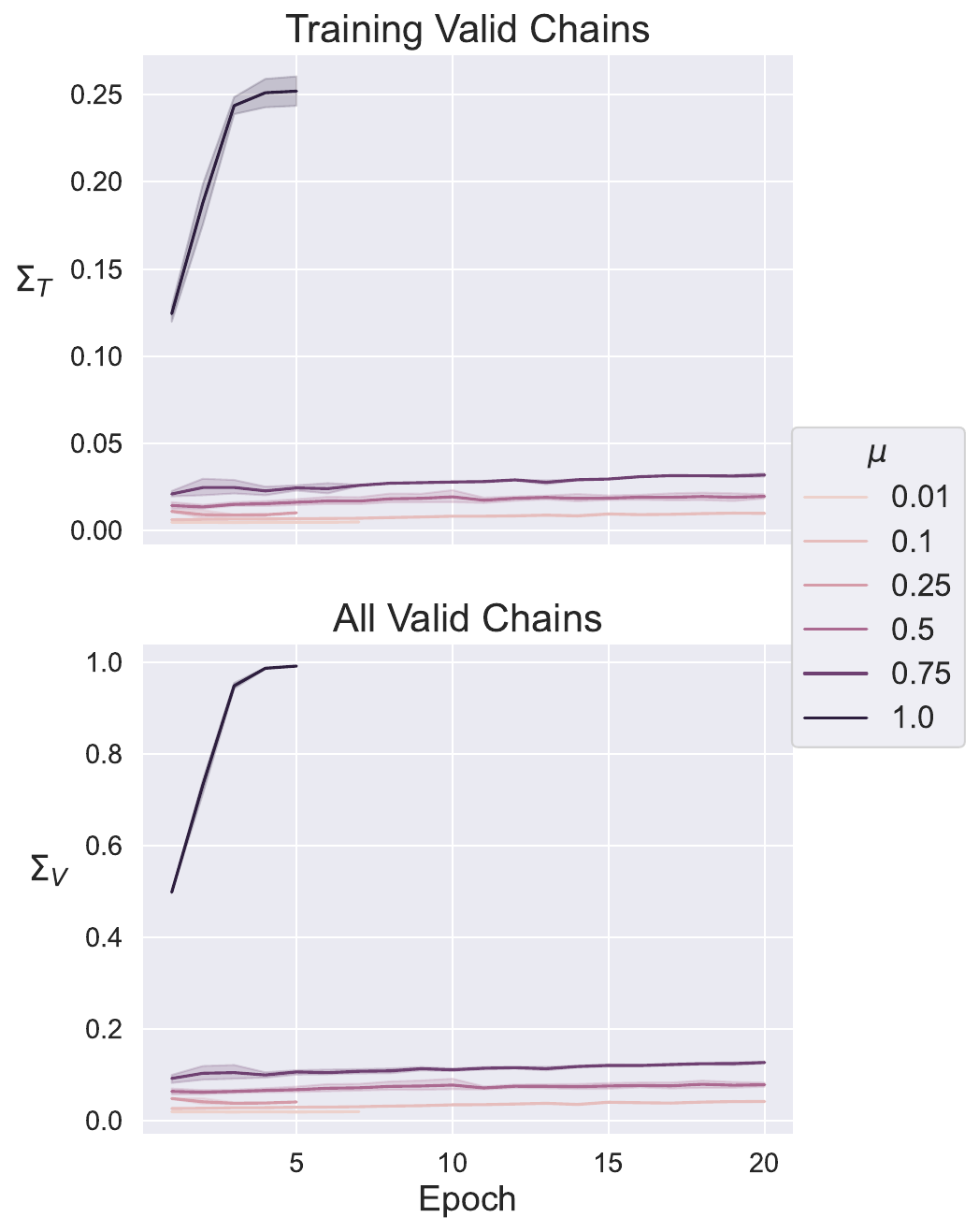}
    \caption{\label{fig:dense_mu_sweep}Dense core probability masses with 95\% confidence interval for various $\mu$.}
\end{figure}
In Figure (\ref{fig:dense_mu_sweep}), we see that the model is able to successfully learn the distribution of the valid Motzkin chain for $\mu=1.0$, but not for lower $\mu$.
As with Tables (\ref{tab:dense_chi}) and (\ref{tab:fact_chi_sweep}), despite low $\Sigma_V$ performance, the model continues to perform well with classification in Table (\ref{tab:mu}), except for $\mu=0.01$.

\begin{figure}
    \centering
    \includegraphics[width=\linewidth]{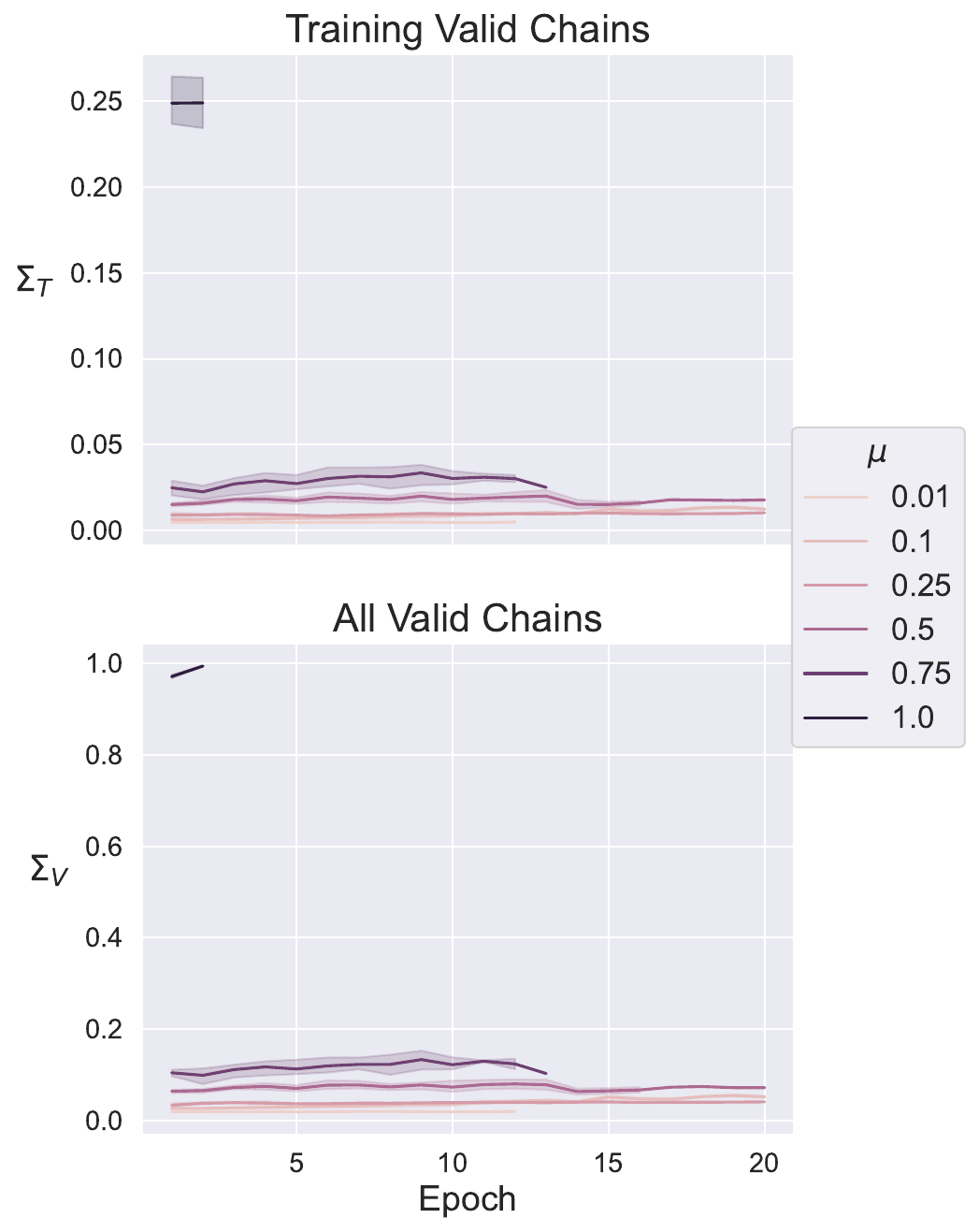}
    \caption{\label{fig:fact_mu_sweep}Factored core probability masses with 95\% confidence interval for various $\mu$, without skip connection.}
\end{figure}

\begin{figure}
    \centering
    \includegraphics[width=\linewidth]{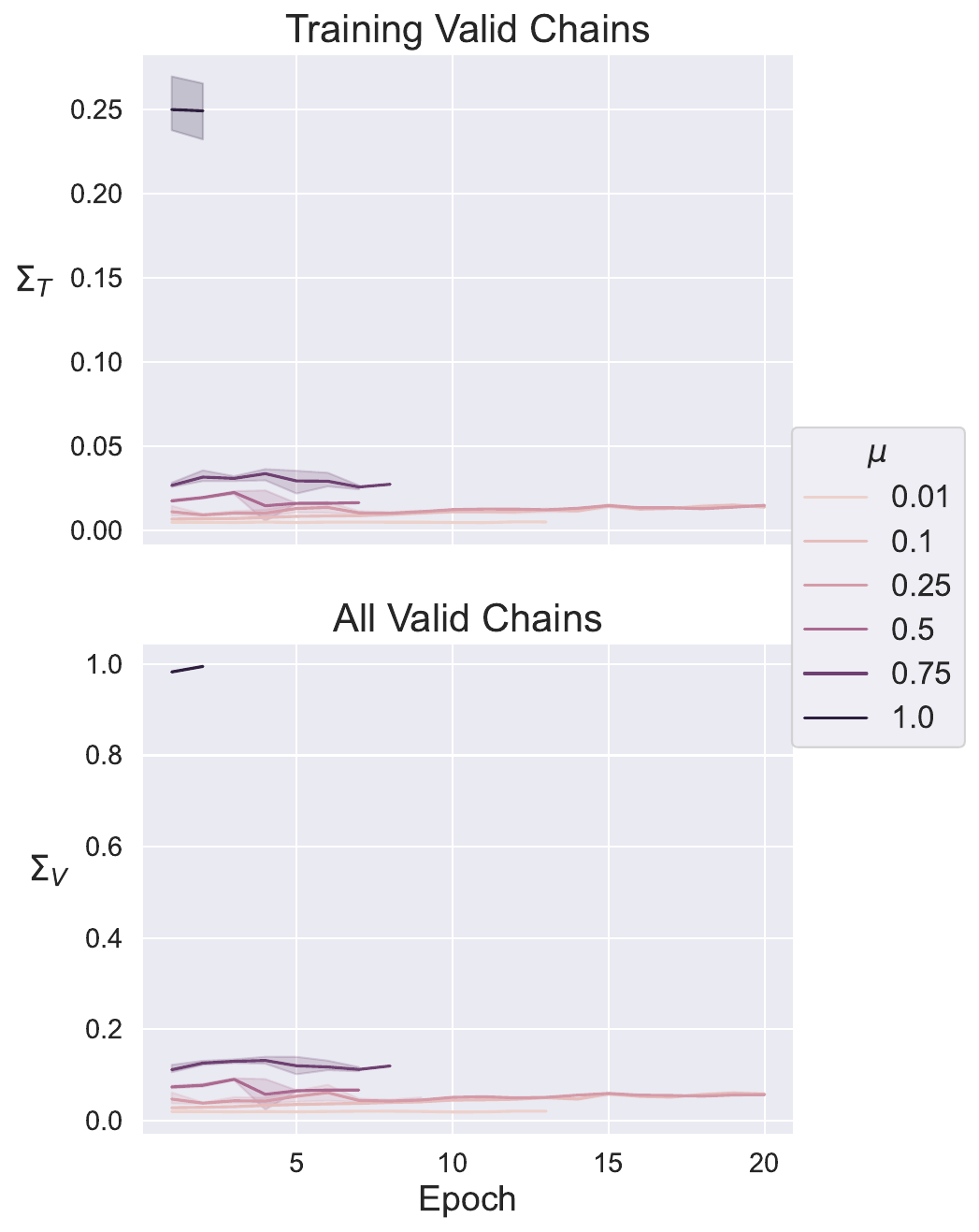}
    \caption{\label{fig:skip_mu_sweep}Factored core probability masses with 95\% confidence interval for various $\mu$, with skip connection.}
\end{figure}

The factored core model's results without skip connection are shown in Figure (\ref{fig:fact_mu_sweep}), and with skip connection in Figure (\ref{fig:skip_mu_sweep}).
We see similar results as with the 10 seed classification experiment.
In both cases, the model that does learn, $\mu=1.0$, does so with fewer epochs of iteration over the training data than the dense core model.
As with the dense core model, for the lower values of $\mu$, the model is unable to capture the distribution over the valid Motzkin chains.
The models are able to function as a good classifier, except when trained with $\mu=0.01$.

\subsection{Comparison of the Tensor Networks' Robustnesses}
\begin{figure*}
    \centering
    \includegraphics[width=0.75\textwidth]{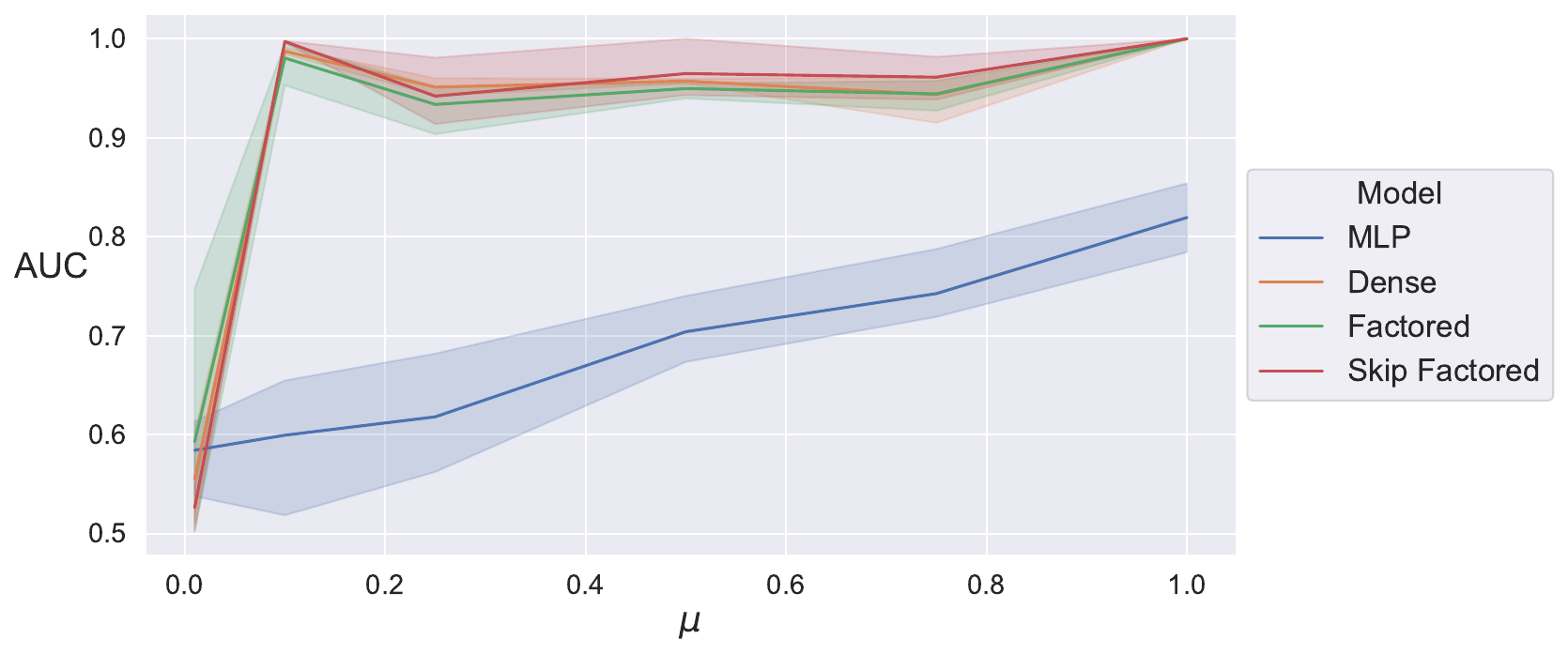}
    \caption{\label{fig:mu_sweep}ROC AUC vs. $\mu$ with 95\% CI for all models, with $\text{seed}=1$ removed.}
\end{figure*}

In Figure (\ref{fig:mu_sweep}), we display all three model's robustness performance, in terms of AUC vs $\mu$, with 95\% confidence interval bands.
Note that we have removed the runs corresponding to the seed with value 1, since they led to an outlier result.

We see that the neural model has consistently mediocre performance.
The tensor network models, in contrast, have generally good performance, that falls off for small $\mu$.
In this case, small $\mu$ mean the fraction of valid chains in the training dataset is smaller than the fraction of valid chains out of all possible chains ($\approx0.0198$), for the given sequence length ($n=16$).
In the low $\mu$ regime, the models perform very similarly, but the tensor network models perform consistently better in other cases.

\section{Discussion}
We have examined using tensor networks to describe Motzkin spin chains, a one-dimensional dataset with long range correlations.
While examining the tensor network's sensitivity to hyperparameters, we have used the Multi-Layer Perceptron (MLP) to ground our analysis in the familiar.
Due to their origin in quantum physical theories, tensor networks offer a novel method of interpreting results, supported by an established theoretical ecosystem.

We have seen that the tensor models can train remarkably well as a classifier with very low quantities of valid chain examples.
This can be seen despite the fact that the model fails to capture all of the probability mass ($\mu=0.75$ and below).
This seems unintuitive: how can a model classify if it doesn't understand the underlying distribution?
It is important to note here that binary classification only really depends on the difference between probabilities.
As such, it is possible that the probability mass for invalid chains is smeared thin enough that it is still substantially lower than the probability mass attributed to the valid chains, making effective classification possible.

It is worth pointing out that, when $\mu$ drops to 0.01, the tensor models barely do better than a random classifier.
This is the regime where our neural baseline pulls ahead, suggesting that neither model-type is a clear winner.
Instead, it behooves a machine learning practitioner to consider each as a tool that has greatest effect when played to its strengths.

\section{Future Work}
Two results emerged that are still rather surprising at the time of writing this paper.
Firstly, an intriguing phenomenon is the batch size dependence of the tensor models.
Specifically, that they perform significantly better with smaller batch sizes, instead of larger ones (to the point of the models' inability to learn at all with large batch sizes).
This an important caveat, since parallelization techniques allow larger batch sizes to be computed more quickly, reducing wall time of training.
We initially thought that, since tensor networks have a vector norm, see Sec. (\ref{sec:mps_norm}), and neural networks do not, this difference may be part of the explanation.
By testing the performance of the tensor network with an alternate norm calculation (eschewing the normalization step in the training loss calculation, or using a simple $l_2$ norm instead), we rule out this characteristic as a possible explanation.
The cause of this batch size dependence is an exciting area for future work.

Secondly, we find that the neural model functions better with a training dataset based purely on valid chains, as opposed to a mix, for it to be able to compare and contrast.
There is literature that suggests that neural models perform fairly consistently across training dataset skewness \cite{Larasati_2019}.
A more thorough investigation of the threshold where tensor network performance drops as a function of $\mu$ is also suitable for future work.

\section*{Limitations}
Instead of being able to use off-the-shelf model implementations, via e.g. HuggingFace, all models had to be built from scratch.
This required several months of debugging and testing, to validate that the models were working as intended.
This was time that could have been used for further experimentation.

Additionally, the limitations in time also led to the decision to use a smaller, toy dataset.
Being a novel combination of model and task (to the author's knowledge), care was taken to examine the effects of the models' various hyperparameters.
With a full natural language dataset, the time needed to perform hyperparameter sweeps would have been greatly increased, as well as adding additional complexity for the aforementioned debugging and testing.

\section*{Ethics Statement}
Given that the data, the Motzkin spin chains, reside in the realm of mathematical objects, there were no considerations needed regarding fairness or representation.
However, great care was taken in making the results reproducible, down to releasing the full source code to an open GitHub repository at \url{https://github.com/ConstantijnvdP/eidolon}.
The goal is not only to reduce the barrier to participate in this line of research, but also to encourage others to question and reproduce the results.
This represents the research ethics credo used to write this paper: that reproducibility and healthy skepticism underlie scientific integrity.

\section{Acknowledgments}
We'd like to thank \href{https://www.sandboxaq.com/}{SandboxAQ} for access to cloud compute resources used in a portion of this work.

This paper built on the thesis work performed by Constantijn van der Poel under advisement from David Schwab and John Terilla of the City University of New York.

\bibliography{paper}

\begin{thebibliography}{20}
\expandafter\ifx\csname natexlab\endcsname\relax\def\natexlab#1{#1}\fi

\bibitem[{Alexander et~al.(2021)Alexander, Evenbly, and Klich}]{Alexander_2021}
Rafael~N. Alexander, Glen Evenbly, and Israel Klich. 2021.
\newblock \href {https://doi.org/10.22331/q-2021-09-21-546} {Exact holographic tensor networks for the motzkin spin chain}.
\newblock \emph{Quantum}, 5:546.

\bibitem[{Bradley(2020)}]{tai_thesis}
Tai-Danae Bradley. 2020.
\newblock \href {https://doi.org/10.48550/ARXIV.2004.05631} {At the interface of algebra and statistics}.

\bibitem[{Bradley et~al.(2020)Bradley, Stoudenmire, and Terilla}]{Qstates_seq}
Tai-Danae Bradley, E~M Stoudenmire, and John Terilla. 2020.
\newblock \href {https://doi.org/10.1088/2632-2153/ab8731} {Modeling sequences with quantum states: a look under the hood}.
\newblock \emph{Machine Learning: Science and Technology}, 1(3):035008.

\bibitem[{Bravyi et~al.(2012)Bravyi, Caha, Movassagh, Nagaj, and Shor}]{crit_spin1_chain}
Sergey Bravyi, Libor Caha, Ramis Movassagh, Daniel Nagaj, and Peter~W. Shor. 2012.
\newblock \href {https://doi.org/10.1103/physrevlett.109.207202} {Criticality without frustration for quantum spin-1 chains}.
\newblock \emph{Physical Review Letters}, 109(20).

\bibitem[{Commons(2021)}]{motzkin_pic}
Wikimedia Commons.
\newblock \href {https://upload.wikimedia.org/wikipedia/commons/b/b7/Motzkin4.svg} {An interpretation of motzkin numbers, the 9 paths from (0, 0) to (4, 0) using only steps northeast, east, and southeast, never dipping below the y-axis.} [online]. 2021.

\bibitem[{Donaghey and Shapiro(1977)}]{motzkin_numbers}
Robert Donaghey and Louis~W Shapiro. 1977.
\newblock \href {https://doi.org/https://doi.org/10.1016/0097-3165(77)90020-6} {Motzkin numbers}.
\newblock \emph{Journal of Combinatorial Theory, Series A}, 23(3):291--301.

\bibitem[{Hornik et~al.(1989)Hornik, Stinchcombe, and White}]{mlp_uni_approx}
Kurt Hornik, Maxwell Stinchcombe, and Halbert White. 1989.
\newblock \href {https://doi.org/https://doi.org/10.1016/0893-6080(89)90020-8} {Multilayer feedforward networks are universal approximators}.
\newblock \emph{Neural Networks}, 2(5):359--366.

\bibitem[{Larasati et~al.(2019)Larasati, Hajji, and Dwiastuti}]{Larasati_2019}
A~Larasati, A~M Hajji, and Anik Dwiastuti. 2019.
\newblock \href {https://doi.org/10.1088/1757-899X/523/1/012070} {The relationship between data skewness and accuracy of artificial neural network predictive model}.
\newblock \emph{IOP Conference Series: Materials Science and Engineering}, 523(1):012070.

\bibitem[{Lin and Tegmark(2017)}]{crit_formal_lang}
Henry~W. Lin and Max Tegmark. 2017.
\newblock \href {https://doi.org/10.3390/e19070299} {Critical behavior in physics and probabilistic formal languages}.
\newblock \emph{Entropy}, 19(7).

\bibitem[{Liu et~al.(2019)Liu, Ran, Wittek, Peng, Garc{\'{\i}}a, Su, and Lewenstein}]{Liu_2019}
Ding Liu, Shi-Ju Ran, Peter Wittek, Cheng Peng, Raul~Bl{\'{a} }zquez Garc{\'{\i}}a, Gang Su, and Maciej Lewenstein. 2019.
\newblock \href {https://doi.org/10.1088/1367-2630/ab31ef} {Machine learning by unitary tensor network of hierarchical tree structure}.
\newblock \emph{New Journal of Physics}, 21(7):073059.

\bibitem[{Martyn et~al.(2020)Martyn, Vidal, Roberts, and Leichenauer}]{guifre}
John Martyn, Guifre Vidal, Chase Roberts, and Stefan Leichenauer. 2020.
\newblock \href {https://doi.org/10.48550/ARXIV.2007.06082} {Entanglement and tensor networks for supervised image classification}.

\bibitem[{Miller et~al.(2020)Miller, Rabusseau, and Terilla}]{Miller_2020}
Jacob Miller, Guillaume Rabusseau, and John Terilla. 2020.
\newblock \href {https://doi.org/10.48550/ARXIV.2003.01039} {Tensor networks for probabilistic sequence modeling}.

\bibitem[{Novikov et~al.(2016)Novikov, Trofimov, and Oseledets}]{exp_mach}
Alexander Novikov, Mikhail Trofimov, and Ivan Oseledets. 2016.
\newblock \href {https://doi.org/10.48550/ARXIV.1605.03795} {Exponential machines}.

\bibitem[{Pestun et~al.(2017)Pestun, Terilla, and Vlassopoulos}]{Pestun2017LanguageAA}
Vasily Pestun, John Terilla, and Yiannis Vlassopoulos. 2017.
\newblock Language as a matrix product state.
\newblock \emph{ArXiv}, abs/1711.01416.

\bibitem[{Schollwöck(2011)}]{dmrg}
Ulrich Schollwöck. 2011.
\newblock \href {https://doi.org/10.1016/j.aop.2010.09.012} {The density-matrix renormalization group in the age of matrix product states}.
\newblock \emph{Annals of Physics}, 326(1):96--192.

\bibitem[{Stokes and Terilla(2019)}]{Stokes_2019}
James Stokes and John Terilla. 2019.
\newblock \href {https://doi.org/10.3390/e21121236} {Probabilistic modeling with matrix product states}.
\newblock \emph{Entropy}, 21(12):1236.

\bibitem[{Stoudenmire and Schwab(2017)}]{schwab_tn}
E.~Miles Stoudenmire and David~J. Schwab. 2017.
\newblock \href {http://arxiv.org/abs/1605.05775} {Supervised learning with quantum-inspired tensor networks}.

\bibitem[{Stoudenmire and Schwab(2016)}]{davidSupTensNet}
Edwin Stoudenmire and David~J Schwab. 2016.
\newblock \href {https://proceedings.neurips.cc/paper/2016/file/5314b9674c86e3f9d1ba25ef9bb32895-Paper.pdf} {Supervised learning with tensor networks}.
\newblock In \emph{Advances in Neural Information Processing Systems}, volume~29. Curran Associates, Inc.

\bibitem[{Tangpanitanon et~al.(2022)Tangpanitanon, Mangkang, Bhadola, Minato, Angelakis, and Chotibut}]{Tangpanitanon_2022}
Jirawat Tangpanitanon, Chanatip Mangkang, Pradeep Bhadola, Yuichiro Minato, Dimitris~G Angelakis, and Thiparat Chotibut. 2022.
\newblock \href {https://doi.org/10.1088/1367-2630/ac6232} {Explainable natural language processing with matrix product states}.
\newblock \emph{New Journal of Physics}, 24(5):053032.

\bibitem[{van~der Poel(2023)}]{my_thesis}
Constantijn van~der Poel. 2023.
\newblock \href {https://academicworks.cuny.edu/gc_etds/5244/} {A quantum approach to language modeling}.

\end{thebibliography}
\bibliographystyle{acl_natbib}

\newpage
\appendix

\section{Overview of Parameters and Their Values}
For quick reference, we include tables of global and model variables, along with their symbols and values in Table (\ref{tab:global_vars}).
\begin{table*}
\centering
\begin{threeparttable}
\begin{tabular}{lcc}\toprule
	Name & Symbol & Value \\\midrule
	Sequence length & $n$ & 16\\
	Vocabulary/Physical Dimension & $\mathcal{V}$ & 3 \\
	Total Dataset Size & $\mathcal{D}$ & \numprint{43046721} \\
	Number Valid Data & $\valdata$ & \numprint{853467} \\
	Training Dataset Size & $\traindata$ & \numprint{213366} \\
        Fraction of $\traindata$ that are valid Motzkin chains & $\mu$ & 1.0$^\dagger$ \\\bottomrule
\end{tabular}
\begin{tablenotes}
	\footnotesize
	\item $^\dagger$ $\mu$ is varied in the robustness experiments of Sec. (\ref{sec:robustness}).
\end{tablenotes}
\caption{\label{tab:global_vars}Global variables.}
\end{threeparttable}
\end{table*}

\begin{table*}
\centering
\begin{tabular}{llccc}\toprule
	MLP & Name & Symbol & Value \\\cmidrule(lr){2-4}
	    & Neural Network Embedding Dimension & $d_e$ & 16 \\
	    & Neural Network Hidden Layer Size & $d_h$ & 256 \\
	\midrule
	Dense & Bond Dimension & $\chi$ & 8 \\
	\midrule
	Factored & Core Type & Symbol & Factored & Skip \\
		 & Height & $h$ & 2 & 3 \\
		 & Horizontal Bond Dimension & $\chi_h$ & 3 & 2 \\
		 & Vertical Bond Dimension & $\chi_v$ & 8 & 4 \\\bottomrule
\end{tabular}
\caption{\label{tab:model_vars}Model variables.}
\end{table*}

\section{Neural Hyperparameter Tuning}
Here, we examine the sensitivity to embedding dimension and hidden layer size.
We run each combination of embedding dimension and hidden layer size for four seeds, and display the average in Table (\ref{tab:mlp_arch}).
\begin{table}
\centering
\begin{tabular}{lccc}\toprule
        & \multicolumn{3}{c}{AUC} \\\cmidrule(lr){2-4}
        $d_e$ & $d_h=128$ & $d_h=256$ & $d_h=512$ \\\midrule
	8  & 0.68(4) & 0.66(6) & 0.64(4) \\
	16 & 0.79(5) & 0.81(4) & 0.77(8) \\
	32 & 0.82(4) & 0.82(9) & 0.82(9) \\\bottomrule
\end{tabular}
\caption{\label{tab:mlp_arch}AUC metrics for MLP architectures.}
\end{table}
We use an embedding dimension of 16, and hidden layer size of 256 in further experiments, to trade off performance with model size. \note{Should I try bigger neural models? Or is that not worth the effort?}

\section{Dense Core Hyperparameters}
\subsection{\label{sec:bond_dim}Bond Dimension}
The bond dimension can be though of intuitively as the size of the \enquote{pipe} for information to flow through between elements of an input sequence.
We are interested to see experimentally how the tensor network models' performance is affected by varying the bond dimension, $\chi$.
\begin{figure}
    \centering
    \includegraphics[width=\linewidth]{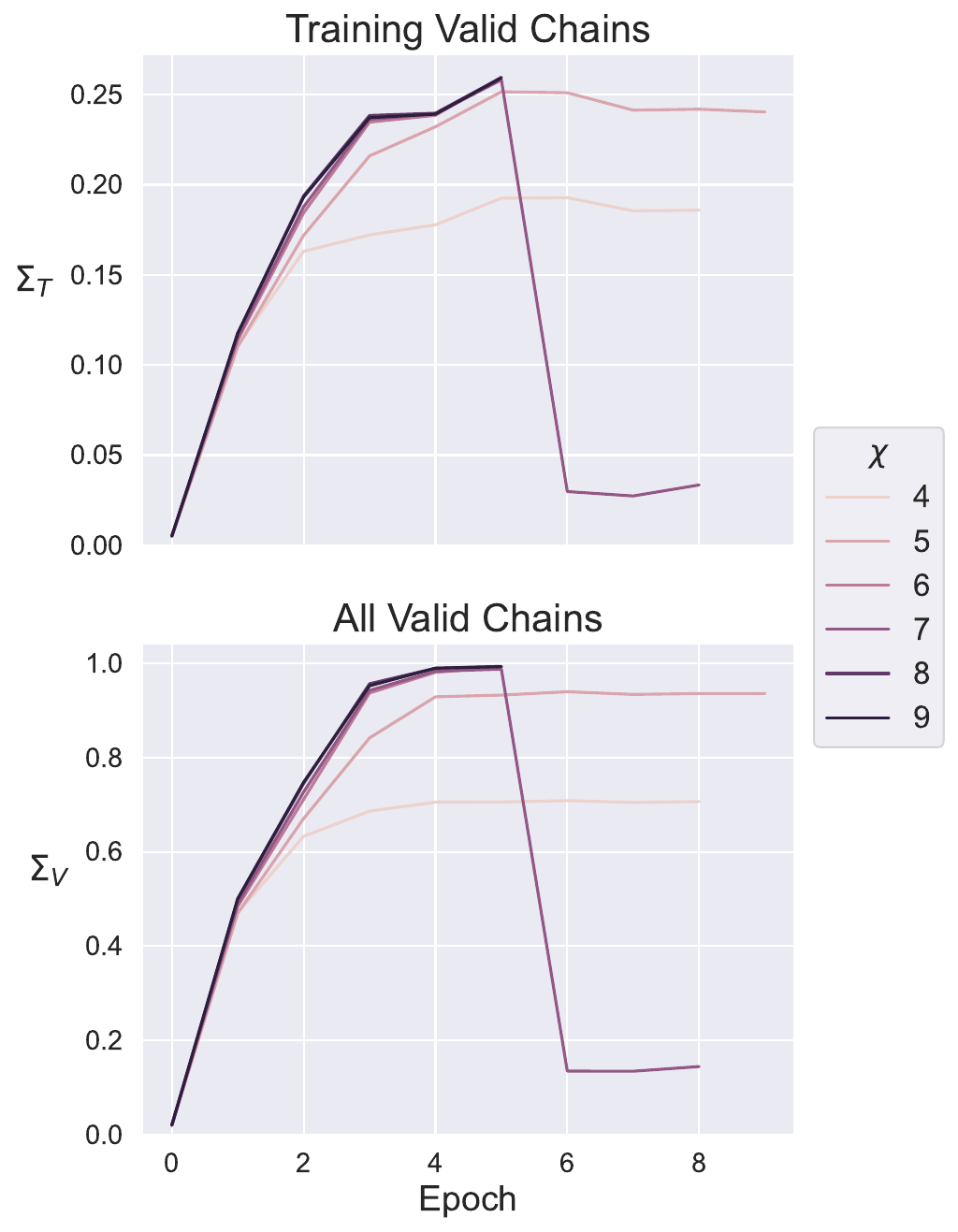}
    \caption{\label{fig:dense_chi}$\chi$ sweep for dense core model.}
\end{figure}
\begin{table}
\centering
\begin{tabular}{lccc}\toprule
	$\chi$ & $\Sigma_T$ & $\Sigma_V$ & AUC \\\midrule
        4 & 0.1859 & 0.7065 & 0.9993 \\
        5 & 0.2404 & 0.9359 & 0.9999 \\
        6 & 0.2598 & 0.9914 & 1.0000 \\
        7 & 0.0334 & 0.1443 & 0.9431 \\
        8 & 0.2595 & 0.9934 & 1.0000 \\
	9 & 0.2593 & 0.9929 & 1.0000 \\\bottomrule
\end{tabular}
\caption{\label{tab:dense_chi}$\chi$ metrics for dense core model.}
\end{table}
Examine Figure (\ref{fig:dense_chi}) and Table (\ref{tab:dense_chi}).
Even despite the catastrophic forgetting seen for $\chi=7$, all of the models do well with the classification task.
We also see that for $\chi=4$, the model saturates around $\Sigma_V\approx0.71$, indicating that the model does not have enough capacity to learn the structure of the Motzkin data.
Again, this does not seem to significantly impact classification performance.

\subsection{\label{sec:alpha_dep}Alpha Dependence}
We are interested in the effect of adding the norm to the loss (Eq. \ref{eq:loss}), which is controlled via an $\alpha$ parameter:
\begin{equation}
	\mathcal{L}_\alpha = \mathcal{L} + \alpha\ln\braket{\psi},
\end{equation}
We tested $\alpha$ values of 0.0, 0.25, 0.5, 0.75, 1.0; each with 4 seeds.
\begin{figure}
    \centering
    \includegraphics[width=\linewidth]{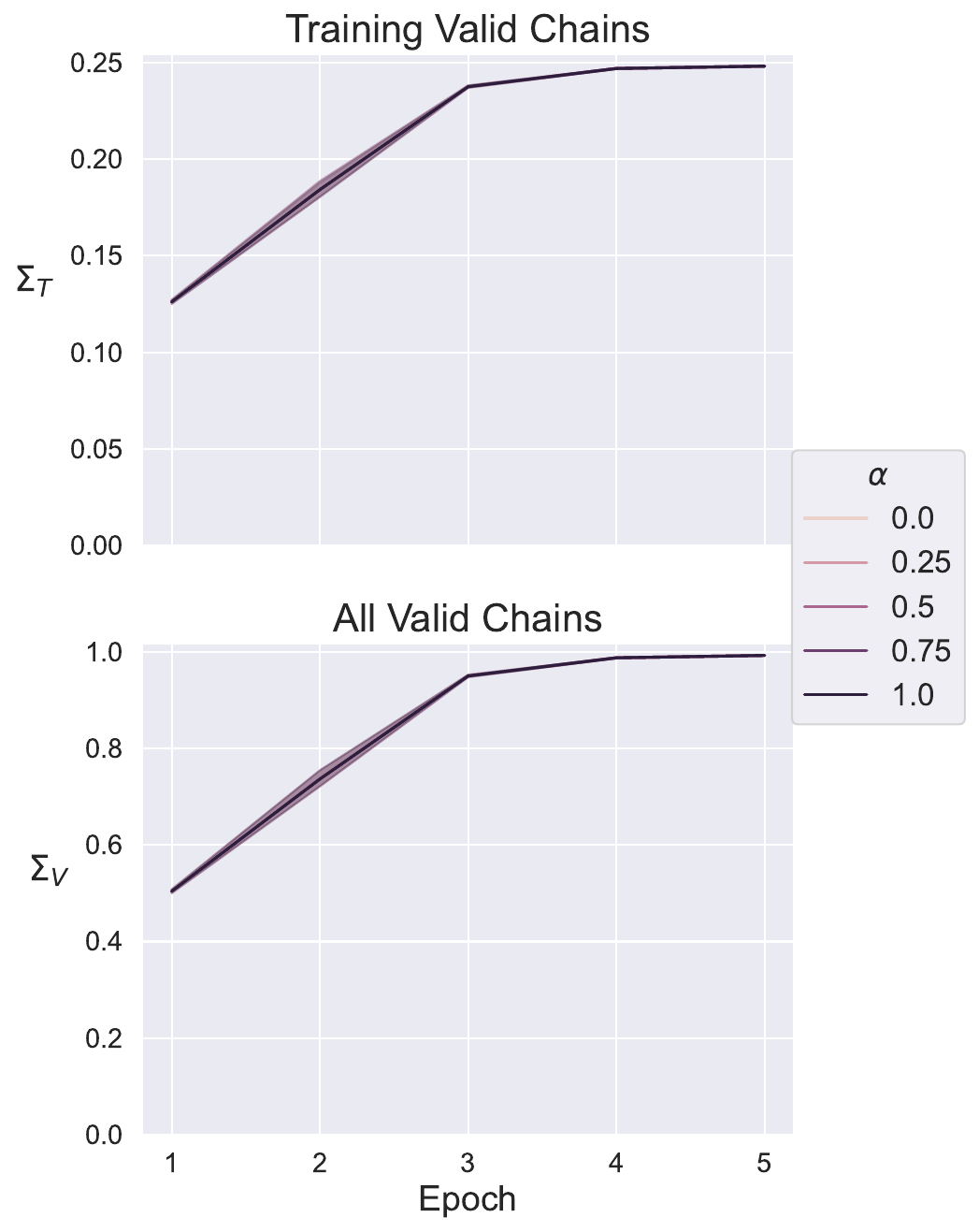}
    \caption{\label{fig:mps_alpha}$\alpha$ sweep for dense core model, with 95\% confidence intervals.}
\end{figure}
We did not see major effects on the model's performance, so we took $\alpha=0$ for all other experiments.

As was discussed in Sec. (\ref{sec:dense_init}), the tensor network cores are initialized to be close to identity.
This is done to facilitate information transfer along the sequence.
If this is indeed conducive to a performant model, then learning will keep the cores near identity.
We speculate this will also serve to keep the norm in check, without needing an additional term in the loss to do so manually.

\subsection{\label{sec:replace_norm}Alternative Norms}
A large part of the computational cost of training a tensor network model lies in its normalization.
We are interested in examining whether this calculation could be replaced by a faster computation.
Furthermore, we are interested to see if the tensor network's norm could be an explanation for its strange batch size dependence.

As a first-order naive attempt, we see what happens when we ignore the norm calculation entirely, assuming it to be 1.0 when calculating the loss.
We perform a sweep of the batch size, taking values 8, 32, 128, 512, and 1024.
\begin{figure}
    \centering
    \includegraphics[width=\linewidth]{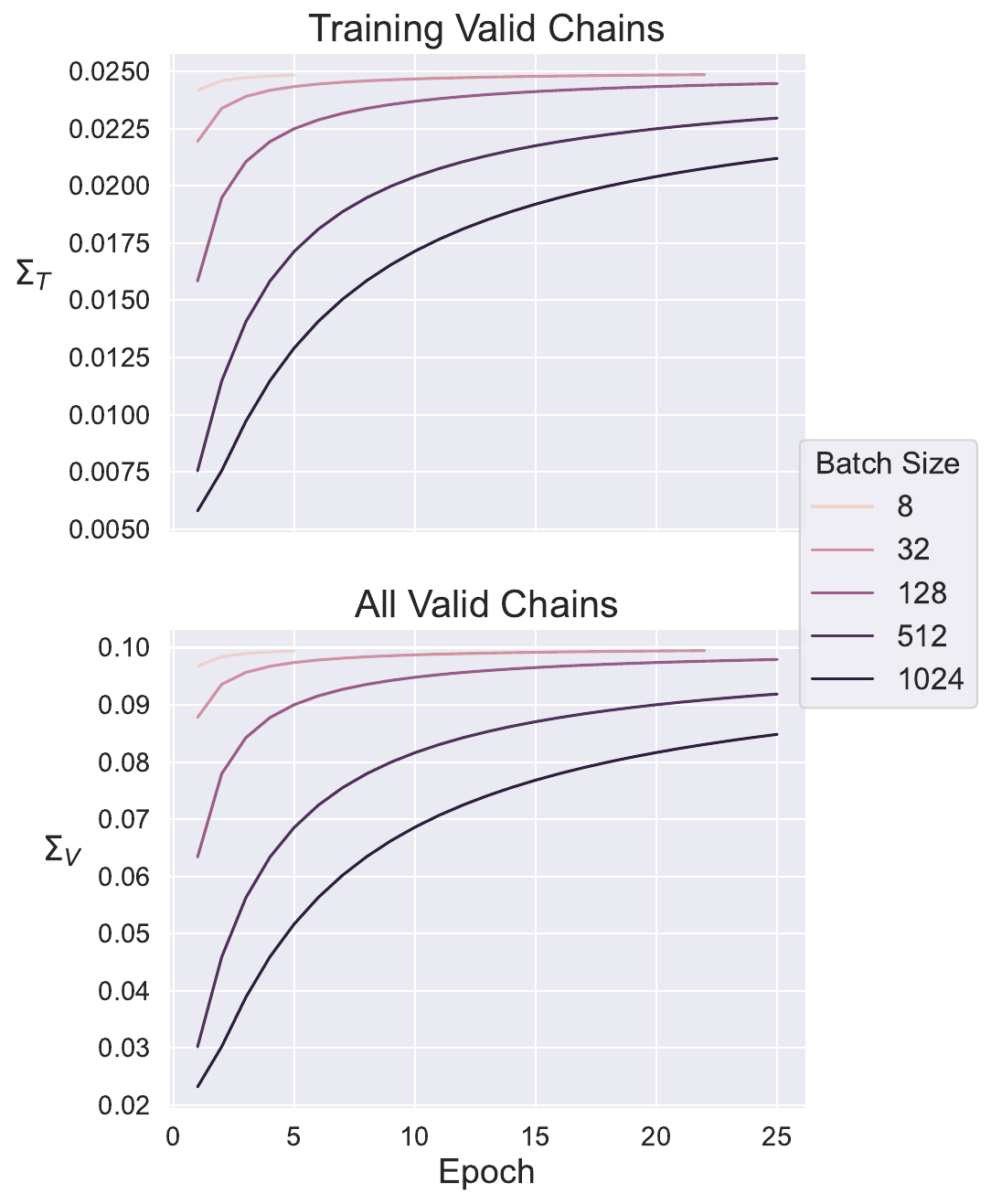}
    \caption{\label{fig:no_mps_norm}Set dense core norm to 1.0, batch size sweep.}
\end{figure}
In Figure (\ref{fig:no_mps_norm}), the models all tend towards the same total valid probability mass.
However, this apparent asymptote is nowhere near the desired full 1.0 probability; the models simply do not learn the data.
As seen previously in Sec. (\ref{sec:bs}), increasing batch size seems to decrease performance here as well.
Because we have removed the norm from the training loss, this cannot be an explanation for this strange behavior.

We will also attempt to replace the norm with the architecture-agnostic $l_2$ norm.
This norm simply sums the square of each parameter, independent of how they are configured.
\begin{figure}
    \centering
    \includegraphics[width=\linewidth]{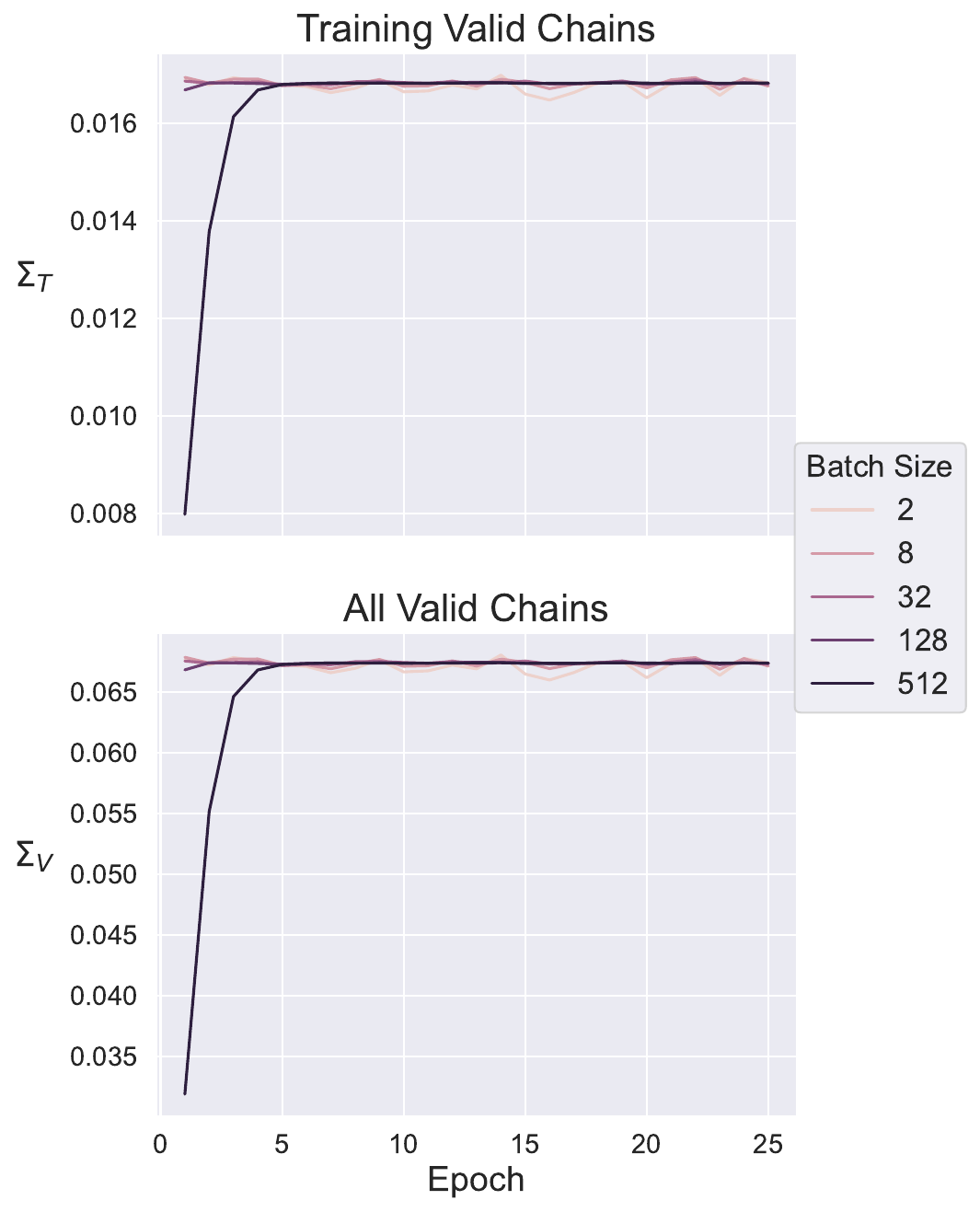}
    \caption{\label{fig:l2_norm}Replace dense core norm with $l_2$ norm, batch size sweep.}
\end{figure}
In Figure (\ref{fig:l2_norm}), we see even less evidence of learning; the norm seems to just throw off the model.

\subsection{\label{sec:dense_init}Initialization Variance}
Let's take a closer look at the variance in the initialization function.
Both inner and outer noise are drawn from normal distributions with mean zero.
With four seeds each, we run our model on the Motzkin dataset of length sixteen.
The results are as follows:
\begin{table}
\small
\centering
\begin{tabular}{lccc}\toprule
	& \multicolumn{3}{c}{$\Sigma_V$} \\\cmidrule(lr){2-4}
	$\sigma_\text{inner}$ & $\sigma_\text{outer}=0.01$ & $\sigma_\text{outer}=0.1$ & $\sigma_\text{outer}=1.0$ \\\midrule
        0.01 & 0.9929 & 0.9930 & 0.7788 \\
        0.1  & 0.9922 & 0.9912 & 0.9914 \\
        1.0  & 0.1465 & 0.1709 & 0.1541 \\\bottomrule
\end{tabular}
\caption{\label{tab:slp_arch}Initialization variance vs. $\Sigma_V$.}
\end{table}
In Table (\ref{tab:slp_arch}), we see that, in general, lower variance performs better.
Also, it is interesting to note that the inner variance is much more correlated with performance than outer.

\section{Factored Core Hyperparameters}
\subsubsection{Factored Core Model}
For the factored cores, this is represented by both the horizontal and vertical bond dimensions, respectively $\chi_h$ and $\chi_v$.
\begin{figure*}
    \centering
    \includegraphics[width=\linewidth]{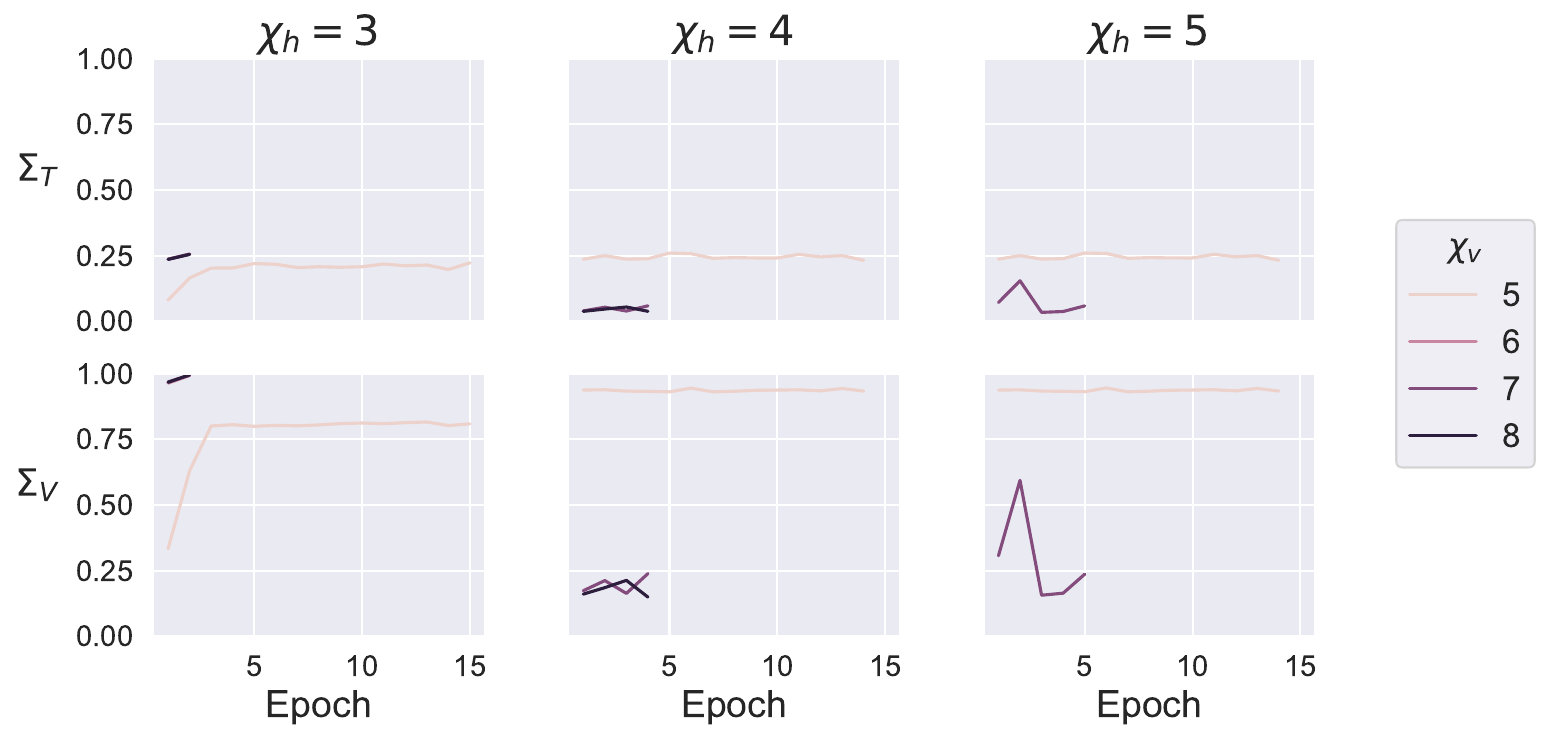}
    \caption{\label{fig:fact_chi_sweep}$\chi_h$ and $\chi_v$ sweep for factored core model.}
\end{figure*}
\begin{table*}
\centering
\begin{tabular}{lccccccccc}\toprule
	& \multicolumn{3}{c}{$\chi_h=3$} & \multicolumn{3}{c}{$\chi_h=4$} & \multicolumn{3}{c}{$\chi_h=5$} \\\cmidrule(lr){2-4}\cmidrule(lr){5-7}\cmidrule(lr){8-10}
	$\chi_v$ & $\Sigma_T$ & $\Sigma_V$ & AUC & $\Sigma_T$ & $\Sigma_V$ & AUC & $\Sigma_T$ & $\Sigma_V$ & AUC \\\midrule
	5 & 0.2226 & 0.8092 & 0.9999 & 0.2320 & 0.9340 & 0.9999 & 0.2320 & 0.9341 & 0.9999 \\
	6 & 0.2559 & 0.9932 & 1.0000 & 0.2420 & 0.9923 & 1.0000 & 0.2428 & 0.9925 & 1.0000 \\
	7 & 0.2556 & 0.9948 & 1.0000 & 0.0589 & 0.2388 & 0.9719 & 0.0585 & 0.2364 & 0.9695 \\
	8 & 0.2547 & 0.9958 & 1.0000 & 0.0378 & 0.1490 & 0.9503 & 0.2419 & 0.9960 & 1.0000 \\\bottomrule
\end{tabular}
\caption{\label{tab:fact_chi_sweep}$\chi_h$ and $\chi_v$ sweep.}
\end{table*}
Interestingly, in Table (\ref{tab:fact_chi_sweep}), we see that the tensor network models' ability to classify chains, as measured by the ROC AUC metric, continues to be strong even when their ability to learn the probability distribution, measured by $\Sigma_V$, is weak, as seen by some combination of bond dimensions in Figure (\ref{fig:fact_chi_sweep}).

\end{document}